\documentclass[11pt,a4paper]{article}
\usepackage{times,latexsym}
\usepackage{url}
\usepackage[T1]{fontenc}

\usepackage{graphicx}
\usepackage{layout}
\usepackage{lipsum}
\usepackage{makecell}
\usepackage{booktabs}
\usepackage{tipa}
\usepackage{hyperref}
\usepackage{cleveref}
\usepackage{float}

\usepackage[acceptedWithA]{tacl2021v1}

\usepackage{xspace,mfirstuc,tabulary}

\newif\iftaclinstructions
\taclinstructionsfalse 
\iftaclinstructions
\renewcommand{\confidential}{}
\renewcommand{\anonsubtext}{(No author info supplied here, for consistency with
TACL-submission anonymization requirements)}
\newcommand{\instr}
\fi

\iftaclpubformat 

\else

\fi

\makeatletter
\newcommand\footnoteref[1]{\protected@xdef\@thefnmark{\ref{#1}}\@footnotemark}
\makeatother

\title{Tracking the emergence of linguistic structure \\in self-supervised models learning from speech}

\author{
  Marianne de Heer Kloots$^{1\Thanks{Corresponding author: m.l.s.deheerkloots@uva.nl}}$,
  Martijn Bentum$^2$,
  Hosein Mohebbi$^3$,\\
  \textbf{Charlotte Pouw$^1$,
  Gaofei Shen$^3$,
  Willem Zuidema$^1$}
  \\[0.5em]
  $^1$Institute for Logic, Language and Computation, University of Amsterdam, The Netherlands
  \\
  $^2$ Centre for Language Studies, Radboud University, The Netherlands
  \\ $^3$ Cognitive Science and Artificial Intelligence, Tilburg University, The Netherlands
}
\date{}

\begin{document}
\maketitle

\begin{abstract}
  Self-supervised speech models learn effective representations of spoken language, which have been shown to reflect various aspects of linguistic structure. But when does such structure emerge in model training? We study the encoding of a wide range of linguistic structures, across layers and intermediate checkpoints of six Wav2Vec2 and HuBERT models trained on spoken Dutch. We find that different levels of linguistic structure show notably distinct layerwise patterns as well as learning trajectories, which can partially be explained by differences in their degree of \emph{abstraction} from the acoustic signal and the \emph{timescale} at which information from the input is integrated. Moreover, we find that the level at which pre-training objectives are defined strongly affects both the layerwise organization and the learning trajectories of linguistic structures, with greater parallelism induced by higher-order prediction tasks (i.e. iteratively refined pseudo-labels).
\end{abstract}

\begin{figure*}
\includegraphics[width=\textwidth]{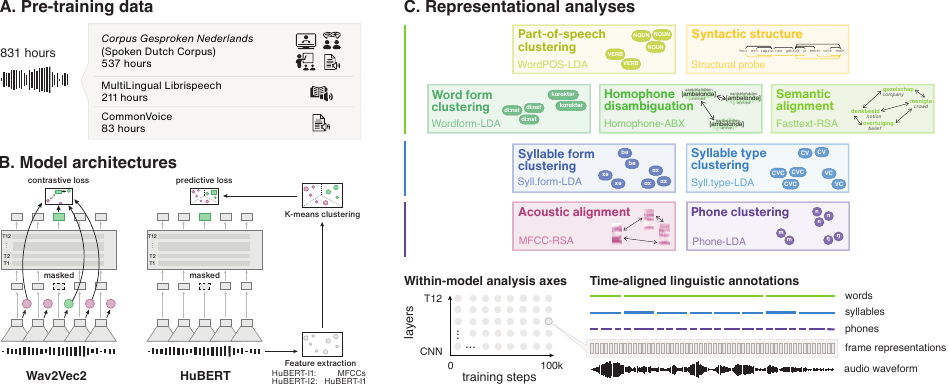}
\vspace{-1em}
\caption{(A) We train a set of six models on the same dataset, consisting of 831 hours of Dutch speech recordings. (B) We explore how results vary between model architectures with minimal differences in training set-up (Wav2Vec2, HuBERT-I1, HuBERT-I2). (C) We probe each model's internal representations for nine types of linguistic structure, which differ in their degrees of abstraction from the acoustic signal and in their timescales of information integration. We compare results for each structure across model layers as well as training steps.}
\label{fig:methods}
\end{figure*}

\section{Introduction}
Much progress in speech technology has been driven by the use of self-supervised learning (SSL) algorithms, which learn powerful representations of spoken language based on unlabelled audio recordings of speech \citep[e.g.][]{baevskiWav2vecFrameworkSelfSupervised2020, hsuHuBERTSelfSupervisedSpeech2021, chenWavLMLargeScaleSelfSupervised2022a, poliSpidRLearningFast2025}. As foundation models for the speech signal, speech SSL model (S3M) representations form a crucial component in many state-of-the-art systems for downstream speech technology tasks, including tasks targeting the signal's \emph{language content}, like automatic speech recognition \citep[e.g.][]{teamOmnilingualASROpenSource2025a} and spoken language modeling \citep{aroraLandscapeSpokenLanguage2025}. 

Current S3Ms are typically Transformer architectures optimized on large collections of recordings with relatively simple pre-training objectives (e.g. masked audio segment prediction; \citealp[see][for review]{mohamedSelfSupervisedSpeechRepresentation2022}). Given these primarily acoustic objectives, it is not obvious that such models should learn to encode more abstract structural information about spoken language. Nevertheless, a substantial amount of work has assessed the linguistic structures encoded in S3M-internal states, often finding that they vastly outperform acoustic baselines when probed for linguistic information, including at the level of phonemes \citep{martinProbingSelfsupervisedSpeech2023}, words \citep{pasadWhatSelfSupervisedSpeech2024} and sentences \citep{shenWaveSyntaxProbing2023a}.

Speech-based SSL models have also been proposed as candidate systems for the modelling of \emph{human} language learning and processing, given their ability to operate on more realistic input signals than text-based models \citep{dupouxCognitiveScienceEra2018}. One challenge that speech poses to both humans and machine learning systems alike is the reliable identification of linguistic units in continuous signals, which themselves vary widely across speakers, contexts, and acoustic conditions (i.e. the \emph{lack of invariance problem}; \citealp{perkellInvarianceVariabilitySpeech1986, UnderstandingSpeechContext2016}). While cognitive theories have posited that a sensitivity for symbolic categories is needed to overcome this challenge \citep[e.g.][]{werkerINFLUENCESINFANTSPEECH1999, kuhlEarlyLanguageAcquisition2004}, S3Ms form an interesting counter-perspective to this idea, providing evidence of what linguistic structures can in fact get represented by systems using purely distributional learning mechanisms while operating on continuous signals alone \citep{schatzEarlyPhoneticLearning2021, lavechinSimulatingEarlyPhonetic2025}.

The linguistic organization of speech importantly follows a hierarchy that can be characterized in terms of both \emph{timescale} and \emph{abstraction}: from short-range acoustic features to long-range and abstract syntactic dependencies. How different levels of linguistic structure might interact in human speech perception has long been a central topic of debate: while some have modelled the process of spoken language understanding as a series of relatively encapsulated prelexical, lexical, syntactic and semantic stages \citep[for discussion, see][]{weberModelsSpokenwordRecognition2012,friedericiNeuralBasisAuditory2002,norris2016prediction}, others emphasize the importance of interactions between structural levels \citep[][]{magnuson2018interaction, marslen-wilsonSentencePerceptionInteractive1975, elmanCognitivePenetrationMechanisms1988}. Similarly, some have characterized the language \emph{learning} process as a series of successive stages whereby infants first acquire the basic units of speech before they are able to access the higher-level combinations of them \citep[][]{kuhlNewViewLanguage2000, werkerPerceptualBeginningsLanguage2018}, while contrasting views emphasize the role of larger patterns being acquired earlier in learning \citep[][]{tomaselloConstructingLanguageUsageBased2003, arnonStartingBigApproach2021}, and the use of higher-level structures in learning to identify the basic units \citep{gleitmanStructuralSourcesVerb1990, feldmanRoleDevelopingLexicon2013}. 

Do the internal representations of self-supervised speech models reflect the hierarchical organization of linguistic structures? And does it characterize the \emph{order} in which the different linguistic structures emerge? We here report on a range of interpretability analyses addressing these questions, by investigating model-internal representations across nine levels of linguistic structure. We study the layerwise distribution of these structures as well as their development across training time, for a new set of Wav2Vec2 and HuBERT models trained on 831 hours of spoken Dutch (\autoref{fig:methods}).

\section{Related work}
\subsection{Layerwise hierarchies and linguistic structure in S3Ms}
\emph{Where} different kinds of acoustic, linguistic, and speaker-related information are best encoded across model hidden layer representations has been studied for a wide range of S3M architectures \citep{pasadLayerWiseAnalysisSelfSupervised2021a, pasadComparativeLayerWiseAnalysis2023, mohamedOrthogonalityIsotropySpeaker2024}, as well as models fine-tuned for speech-to-text transcription \citep{pasadLayerWiseAnalysisSelfSupervised2021a, rollCategorizeEarlyIntegrate2026}. These studies have generally found a common pattern across Transformer architectures: local acoustics and voice characteristics peak early, whereas labels abstracting away from such lower-level features (e.g. phone or word identity) are best decodable from middle-to-late model layers. Moreover, training objectives greatly affect the layerwise organization of speech features: whereas the encoding of linguistic features peaks in middle layers of SSL-trained Wav2Vec2 and sharply drops off afterwards (reflecting a specialization towards the model's acoustic pre-training objective), ASR-finetuned Wav2Vec2 preserves such features into its final layers \citep{pasadLayerWiseAnalysisSelfSupervised2021a}.

\emph{What} kinds of linguistic structures end up encoded in S3M representations has been the focus of a growing complementary body of research, with individual studies typically targeting separate structural levels. Collectively, these studies suggest that S3Ms encode a rich set of linguistic feature structures ranging across multiple timescales and levels of abstraction: from allophonic variation \citep{choiLeveragingAllophonySelfSupervised2025} to phoneme categories \citep{martinProbingSelfsupervisedSpeech2023}, phonotactic constraints \citep{deheerklootsHumanlikeLinguisticBiases2024}, morphological patterns \citep{gauthierEmergentMorphophonologicalRepresentations2025}, lexical information \citep{pasadWhatSelfSupervisedSpeech2024}, and syntactic structure \citep{shenWaveSyntaxProbing2023a}. When probed for sentence grammaticality across a range of morphosyntactic phenomena, S3M representations even sometimes match or outperform ASR models trained with textual supervision, demonstrating that grammatical knowledge can indeed arise purely from speech \citep{heLayerwiseMinimalPair2025}.

However, because most linguistic investigations into S3M representations have so far focussed on individual structural levels, with analysis data and methods varying between studies, it is currently unknown whether fine-grained distinctions between levels of linguistic organization (e.g. phone-, syllable-, word- and sentence-level structures) are reflected in S3M layerwise hierarchies. Additionally, since most studies focus on the encoding of English linguistic information in models pre-trained on English speech recordings \citep[with some notable exceptions:][]{milletSelfsupervisedSpeechModels2022, shen-etal-2024-encoding, delafuenteLayerwiseAnalysisMandarin2024a, dugonjicWhatHasLeBenchmark2024, deheerklootsWhatSelfsupervisedSpeech2025}, it is an open question to what extent S3M encoding of linguistic information at various structural levels generalizes to other languages.

\subsection{Learning dynamics in neural models of (spoken) language}

The process of learning linguistic units and meaning from data has been a primary topic of investigation since early connectionist modelling (e.g. \citealp{mcclellandTRACEModelSpeech1986a, rumelhart1986learning, rogersSemanticCognitionParallel2004}). Current research in computational linguistics has started investigating learning dynamics in modern neural language models. In the text domain, model behaviors across training have been found to mimic aspects of human word and syntax learning \citep{changWordAcquisitionNeural2022, evanson-etal-2023-language}, and measurements of model-internal components can be related to drops in the loss and increased syntactic abilities \citep{chenSuddenDropsLoss2023a}. 

In the speech domain, the acquisition of language-specific phonetic and lexical structure has been studied in smaller recurrent SSL models (e.g. \citealp{poliModelingInitialState2024, lavechinSimulatingEarlyPhonetic2025}), as well as in Transformer-based Wav2Vec2 models trained with audiovisual supervision on corpora of spoken image captions \citep{khorramiComputationalInsightsAcquisition2023}, and in Wav2Vec2 models trained only on speech \citep{orhanEmergencePhonemicSyntactic2026}. Both \citeauthor{khorramiComputationalInsightsAcquisition2023} and \citeauthor{orhanEmergencePhonemicSyntactic2026} report that phonemic distinctions precede lexical ones in model representations across training, with syntactic structure \citep{orhanEmergencePhonemicSyntactic2026} emerging later, while both studies also demonstrate some degree of overlap between learning trajectories. 

Hence, evidence exists that S3Ms learn to encode various levels of linguistic information across training, and that the encodings of different levels of linguistic structure might show distinct patterns across model layers and training trajectories. Nevertheless, both aspects have not yet been systematically studied for a wide range of features (including acoustic, sublexical, lexical and  syntactic structures), while simultaneously exploring effects of architectural variations on both layerwise patterns and learning dynamics. Here, we provide such analyses for a set of six Wav2Vec2 and HuBERT models trained on Dutch.

\section{Methods}

\subsection{Models}
\label{sec:model-training}

\paragraph{Architectures}
To explore effects of architectural variation on model representational structure, we include two representative systems applied widely in current speech SSL research, which vary minimally in their architectural configuration: Wav2Vec2 \citep{baevskiWav2vecFrameworkSelfSupervised2020} and HuBERT \citep{hsuHuBERTSelfSupervisedSpeech2021}; see \autoref{fig:methods}B. For both model architectures we use the same base-sized configuration, consisting of a 7-layer CNN module followed by 12 Transformer layers; all randomly initialized before model training. We train six models in total, with two models (initialized with different random seeds) for each of three variations in training objective: \textbf{Wav2Vec2} models use a contrastive objective with negative examples extracted from the output of its feature encoder, \textbf{HuBERT-I1} \emph{(iteration 1)} models use a pseudo-label prediction objective with pseudo-labels extracted by k-means clustering on acoustic (MFCC) features, and \textbf{HuBERT-I2} \emph{(iteration 2)} models use a pseudo-label prediction objective with pseudo-labels extracted by k-means clustering the hidden layer representations of the iteration 1 models. We train HuBERT-I1 models for 100K training steps; pseudo-labels for pre-training HuBERT-I2 were extracted from the 6th layer of the final HuBERT-I1 checkpoints. Wav2Vec2 and HuBERT-I2 models were both trained for 200K training steps. HuBERT-I2 models are equivalent to the HuBERT architecture most commonly applied for downstream applications; we choose to study both iterations here, to disentangle observed effects of contrastive vs. predictive training objective from effects of iterative refinement (following \citealp{huoIterativeRefinementNot2025}).

\paragraph{Training data, set-up and checkpoints} We use the fairseq library \citep{ottFairseqFastExtensible2019} to train all Wav2Vec2 and HuBERT models on 831 hours of Dutch speech recordings from various domains (combining data from the corpus of spoken Dutch (\emph{Corpus Gesproken Nederlands}; CGN), \citealp{schuurmanCGNAnnotatedCorpus2003}; CommonVoice (CV), \citealp{ardilaCommonVoiceMassivelyMultilingual2020}; and MultiLingual LibriSpeech (MLS), \citealp{pratapMLSLargeScaleMultilingual2020}). We extract 537 hours of audio from CGN, covering both Dutch and Flemish material while excluding telephone speech and sermons (due to low sample rate and poor audio quality). The remaining dataset contains spontaneous conversations and interviews, along with read speech and news broadcasts. The CGN recordings are segmented into phrases, and only segments between 2 and 15 seconds are used as training inputs. From MLS, we include 211 hours of audiobook segments. An additional 83 hours are sourced form CommonVoice, consisting of short read‑aloud sentences. Across the full training set, audio segment durations range from 2 to 20 seconds. We follow the default training recipes for both Wav2Vec2 and HuBERT, only modifying fairseq configurations to allow longer utterance length and per-device batch size. We used 16 Nvidia A100-40GB GPUs for model training. Training for 100k steps took approximately 16 hours for HuBERT-I1 models; training for 200k steps took approximately 50 hours for HuBERT-I2 models, and 100 hours for Wav2Vec2 models. For all models, we save intermediate checkpoints in increasing intervals throughout training (every checkpoint between 1 and 100 steps, every 10th checkpoint between 100 and 1,000 steps, every 100th checkpoint between 1,000 and 10K steps, every 1,000th checkpoint between 10K and 100K steps, and every 10,000th checkpoint between 100K and 200K steps). For our representational analyses in this study, we choose to include every 1000th checkpoint up to 100K steps, after verifying that capable models were trained and training up to this step was stable for all architectures (see \autoref{app:model-validation}; \autoref{fig:loss-curves}, \autoref{tab:wer-results}).

\paragraph{Nonspeech baseline model} 
In addition to our speech-trained models, we include a baseline comparison model in all our analyses, to ensure that the linguistic structuring detected by our representational probes is in fact a consequence of model training on speech data, rather than a general reflection of input acoustics. As our nonspeech baseline, we use a Wav2Vec2 model trained for 100K steps on 900 hours of non-speech acoustic scenes from AudioSet, for which intermediate training checkpoints have been released by \citet{orhanDetectionAlgebraicAuditory2025}. This model was verified by \citeauthor{orhanDetectionAlgebraicAuditory2025} to obtain reasonable performance when fine-tuned on a downstream acoustic validation task (environmental scene classification), and additionally showed some capability for recognizing sequential patterns in non-speech auditory stimuli.

\subsection{Representational analyses}
\label{sec:probe-methods}
We apply a range of analysis techniques targeting different kinds of linguistic structure, probing for model-internal alignment with both categorical (e.g. phone, syllable and word labels) and relational (e.g. continuous acoustic and semantic feature spaces) structures (see \autoref{fig:methods}C). 

\paragraph{Analysis data and linguistic annotations} To construct analysis datasets for each of our probes, we first obtain annotations on the level of phones, syllables and words for a subset of recordings in Multilingual LibriSpeech (MLS), which was not included in model training. Taking advantage of the fact that one male Dutch speaker recorded many audiobooks included in MLS, we restrict the subset to this speaker (\texttt{male\_2450}) only; by only analyzing recordings from a single speaker, we ensure that results are not confounded by speaker imbalance across analysis recordings. We obtain time-aligned phonetic transcriptions on the level of phones and words by using WebMAUS\footnote{\url{https://clarin.phonetik.uni-muenchen.de/BASWebServices/interface/WebMAUSBasic}} to force-align the audio files and text transcriptions from MLS. For a subset of words we additionally extract syllable segmentations from the CELEX database \citep{baayenCELEXLexicalDatabase1996}. Finally, for our syntactic analyses (probing for part-of-speech categories and dependency structure), we manually identify a set of 2406 individual sentences (since recordings in MLS are not segmented by sentence), for which we extract part-of-speech (POS) labels and dependency annotations using spaCy \citep{Honnibal_spaCy_Industrial-strength_Natural_2020}; we refer to these annotations, automatically obtained from the sentence text transcriptions, as the `true' POS labels and dependency parses\footnote{We use pipelines based on \href{https://spacy.io/models/nl\#nl\_core\_news\_lg}{nl\_core\_news\_lg}, for which the spaCy documentation reports 95\% POS tagging accuracy and 88\% unlabeled attachment score for dependency parsing.}.

\paragraph{Embedding extraction} All models included in our study generate representations at a framerate of 20 ms. We extract representations for each linguistic unit (phone, syllable or word) by \emph{feature-slicing}, i.e. feeding a full audiobook segment from MLS as input and isolating hidden-state activations within the start and end of each unit. Following common practice 
\citep[e.g.][]{choiSelfSupervisedSpeechRepresentations2024a, pasadWhatSelfSupervisedSpeech2024}, we subsequently mean-pool frame representations within each linguistic unit to obtain phone-, syllable- and word-level embeddings. We analyze each model's 512-dimensional CNN output, as well as the 768-dimensional projections to Transformer embedding space (embeds) and all Transformer hidden layers (T1-12).

\paragraph{Analysis methods} 
We aim to measure the \emph{linguistic structuring} of model-internal representations, but S3M representations are known to entangle many different aspects of the speech signal (including e.g. speaker voice and acoustic characteristics as well as multiple linguistic features). How can we quantify the encoding of specific linguistic information related to our target structures, in such highly entangled representations? One commonly applied method for interpreting model hidden state representations involves the use of diagnostic classification probes, i.e. auxiliary classifiers trained to decode target information from model hidden layer states \citep{hupkesVisualisationDiagnosticClassifiers2018,chrupala-etal-2020-analyzing}. However, classification probes have been considered suboptimal for analyzing model learning dynamics, with alternative \emph{representation space} metrics showing more sensitivity to subtle differences between checkpoints and better alignment to changes in model behavioral performance over training \citep{saphraUnderstandingLearningDynamics2019}. We here use a related range of metrics, which similarly target model representation space structure (i.e. measuring relative distances between input stimuli, clustering, and alignment to interpretable feature spaces), rather than classification accuracy. We analyze a total of nine levels of linguistic structure (\autoref{fig:methods}C), choosing our analysis techniques to suit each categorical and relational linguistic feature structure of interest. 

For all probes analyzing categorical linguistic information (i.e. phone categories, syllable forms and types, word forms and part-of-speech labels), we use \emph{clustering probes} based on dimensionality-reduced subspaces extracted using \textbf{Linear Discriminant Analysis} (LDA; \citealp{tharwatLinearDiscriminantAnalysis2017}). We evaluate model-representational clustering according to target category labels by computing the \emph{silhouette score} on LDA projections optimized for distinctiveness between categories; silhouette scores range between -1 and 1, where 0 corresponds to random structure and more positive scores indicate more well-separated clusters. Details on all clustering analysis datasets are included in \autoref{app:probe-details}; Tables \ref{tab:clustering-details}, \ref{tab:word-data-stats}, \ref{tab:pos-data-stats}, \ref{tab:phone-set}, \ref{tab:syllable-set}.

For analyzing alignment to continuous feature spaces (mel-frequency cepstral coefficients (MFCCs) for acoustic, and static text-based word embeddings for semantic alignment), we use \textbf{Representational Similarity Analysis} (RSA; \citealp{kriegeskorteRepresentationalSimilarityAnalysis2008a}). RSA scores are Pearson's correlation coefficients between the cosine-dissimilarity spaces extracted from model representation and interpretable feature space, respectively; more positive \emph{pearson's r} values indicate higher similarity. For our analyses of \emph{semantic alignment}, we compute similarity to a set of embeddings from a text-based semantic embedding model (Dutch Fasttext\footnote{Accessed from the HuggingFace Hub \citep{wolfHuggingFacesTransformersStateoftheart2020}; \href{https://huggingface.co/facebook/fasttext-nl-vectors}{\texttt{facebook/fasttext-nl-vectors}}}), for which high alignment with human semantic similarity judgments has been reported \cite{bransMultiSimLex2026}. To avoid confounding effects of part-of-speech on our semantic alignment measure, we run analyses separately within part-of-speech category subsets, and to avoid confounding effects of word identity, we exclude distances between samples of the same word. Further details on feature spaces and analysis subsets are included in \autoref{app:probe-details}; \autoref{tab:rsa-features}.

To evaluate model-internal disambiguation between \emph{homophones}, i.e., word pairs with identical pronunciations but different meanings, we construct an \textbf{ABX} task \citep{schatzABXDiscriminabilityMeasuresApplications2016}, where word triplets A, B and X are identical in pronunciation (i.e. in their phonetic transcription), but only A and X are also identical in meaning (i.e. in their orthographic transcription\footnote{Many Dutch homophone pairs, including the ones in our sample, are pairs of present- vs. past-tense verb forms (e.g. \textipa{[amb@lAnd@]} \emph{aanbelanden} / \emph{aanbelandden}) or singular vs. plural verb or noun forms (e.g. \textipa{[briGad@]} \emph{brigade} / \emph{brigaden}).}). Our measure of \emph{homophone disambiguation} is the binary ABX accuracy on each triplet (1 if AX-similarity $>$ AB-similarity, 0 otherwise); chance accuracy is 0.5.

Finally, we use a \textbf{structural probe} paradigm \citep{hewittStructuralProbeFinding2019} to investigate how well model-internal representations encode \emph{syntactic dependency structure} between words. Structural probes were originally introduced for text-based models, and evaluate to what extent linear projections from model embedding space can be trained to reflect distances between words in a sentence that align with distances in that sentence's syntactic dependency parse. We closely follow the original implementation of the structural probe by \citealp{hewittStructuralProbeFinding2019}, replacing the token-level hidden-state activations with our word-level speech model embeddings. Encoding of syntactic structure is evaluated by the Undirected Unlabeled Attachment Score (UUAS; the percentage of correctly placed undirected edges), computed between dependency structures reconstructed from speech model representations, and the set of spaCy-annotated true parses. We confirm that our UUAS score measurements reflect some degree of syntactic and not just linear-sequential structure, by comparing results for sequential dependency structures where each word is only linked to its sequential neighbors instead of its syntactic dependants (\autoref{app:probe-details}, \autoref{fig:syntactic-control-results}).

Variants of all our probing techniques have been applied for interpreting speech model representations before (e.g. LDA and RSA in \citealp{bentumWordStressSelfsupervised2025, sauterCuriousCaseVisual2026}; lexical ABX in \citealp{algayresDPParseFindingWord2022}; structural probing in \citealp{dugonjicWhatHasLeBenchmark2024, orhanEmergencePhonemicSyntactic2026}). All techniques except RSA and ABX involve optimizing auxiliary models to decode target information from model representations; in these cases we evaluate on held-out test sets, designing train-test splits to avoid confounds, and using 5-fold cross-validation. For other cases we simply repeat analyses over 5 folds of data to get an estimate of the variance in scores. We include more details on train-test splits and other probing hyperparameters in \autoref{app:probe-details}. 

\begin{figure}
    \centering
    \includegraphics[width=\columnwidth]{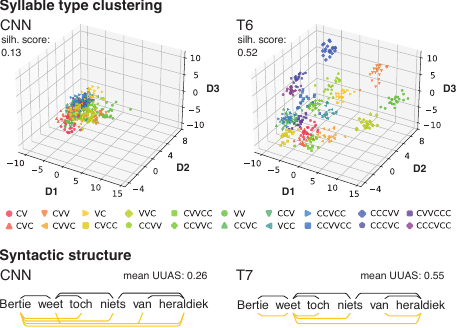}
    \vspace{-1em}
    \caption{Example outputs for two probing techniques applied to different layers of a Wav2Vec2 model at 100K training steps, with accompanying scores. Data points in the top figures visualize the top 3 LDA directions for syllable type test samples, colored by syllable type. Yellow lines in the bottom row visualize model-reconstrcuted dependency links for one test sentence; black lines indicate the true dependency structure.}
    \label{fig:probe-examples}
\end{figure}

\section{Results}
\subsection{\emph{Where} are different levels of linguistic structure encoded?}
We start by examining the layerwise results of our analyses, including only model checkpoints saved at 100K training steps. We verify that our analysis scores capture meaningful layerwise differences in model-internal linguistic structuring: \autoref{fig:probe-examples} shows qualitative examples for two probing techniques and accompanying layerwise scores computed over all test samples. \autoref{fig:layerwise} shows the layerwise probe scores for three models representing the variation in training objectives (results for the other three models are very similar, and included in \autoref{fig:layerwise-seed1}). We generally observe that all analyzed levels of Dutch linguistic structure are better represented in the models trained on Dutch speech as compared to the nonspeech baseline model. 

For Wav2Vec2 and HuBERT-I1, earlier model layers show a sequential pattern of peaks at the level of acoustic, phonetic, and syllabic structure, while lexical and syntactic information is mostly concentrated in the same layer (T7). The relative richness of  middle model layers aligns with observations in other, English-trained S3Ms, with the drop at final model layers indicating these layers' specialization for acoustically oriented training objectives \citep[e.g.][]{huoIterativeRefinementNot2025}.

HuBERT-I2 shows a markedly different pattern: training on higher-layer 
I1 pseudo-labels drives the model's final layers to maintain the higher-level linguistic structures encoded in I1's sixth-layer representations. Interestingly, more abstract structures (beyond word form) are still represented, but now peak before the phonetic, syllable and word form peaks, dropping off again in the final layer. We further note that for all linguistic probes beyond acoustics, HuBERT-I2's peak scores exceed those of HuBERT-I1, even though HuBERT-I1's sixth layer representations are used to derive the prediction targets for HuBERT-I2. This indicates that the higher-level pseudo-labels employed for HuBERT-I2 training (extracted from hidden layers which abstract away from local acoustics, rather than from MFCCs), do stimulate the learning of higher-level linguistic structures, even when such structures are imperfectly encoded in the I1 representations used for pseudo-label extraction (e.g. for the part-of-speech and syntactic structure probes, speech-trained HuBERT-I1 does not outperform the nonspeech baseline model at layer T6).

\subsection{\emph{When} do different levels of linguistic structure become encoded?}
\begin{figure}[H]
    \includegraphics[width=\columnwidth]{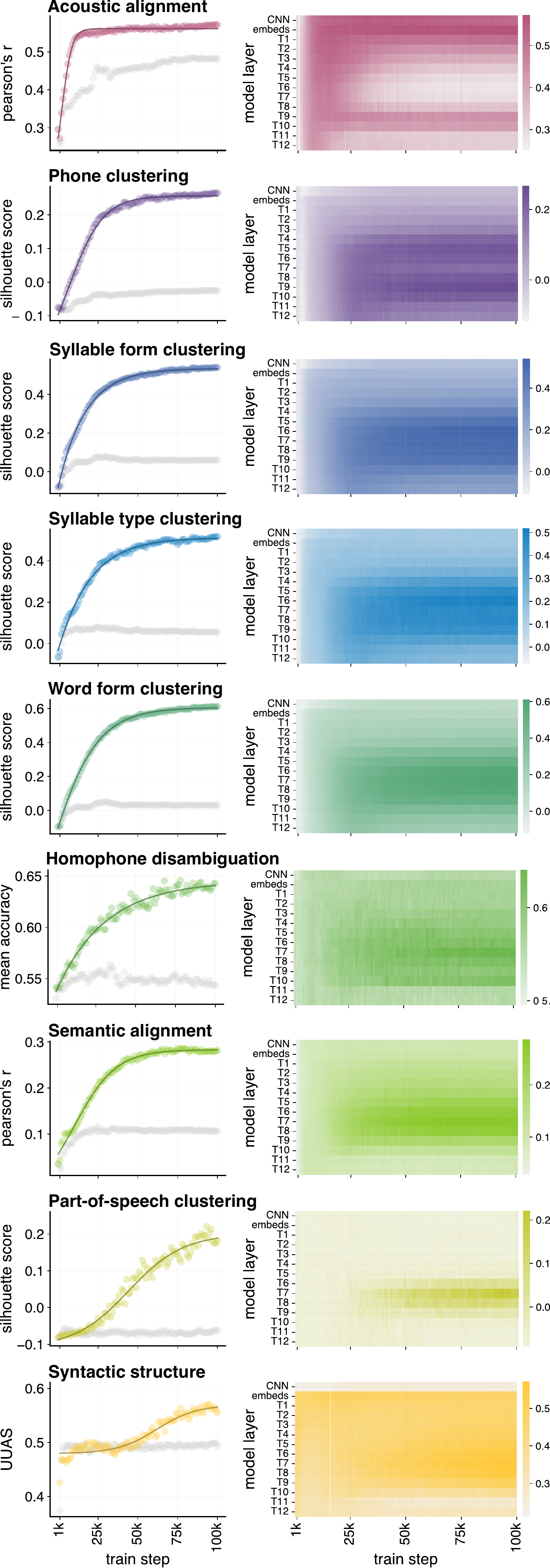}
\vspace{-1em}
\caption{Learning trajectories for all representational probes, across training checkpoints of one Wav2Vec2 model (seed 2). Left: best-layer score for each checkpoint, with fitted sigmoid curves for the speech-trained model; grey dots indicate non-speech baseline model scores. Right: all layerwise scores across training.}
\label{fig:learning-w2v2}
\end{figure}

The \emph{learning trajectories} (i.e. the evolution of probe scores across training checkpoints) for all structures and one Wav2Vec2 model are visualized in \autoref{fig:learning-w2v2}. Observing the general pattern across probes, we note that the encoding of most linguistic structures starts increasing right from the start of model training. The earliest stages of model training seem characterized by learning generally useful representations of speech
acoustics: acoustic alignment scores are the first to reach their maximum performance across training, showing a sharp increase in the first 10k training steps, and remaining relatively stable afterwards. Around 10k steps, most linguistic probe results for the speech-trained model start outperforming those of the nonspeech baseline. The syntactic probes (analyzing the encoding of POS categories and syntactic dependency structures) form an exception to

\begin{figure}[H]
\includegraphics[width=\columnwidth]{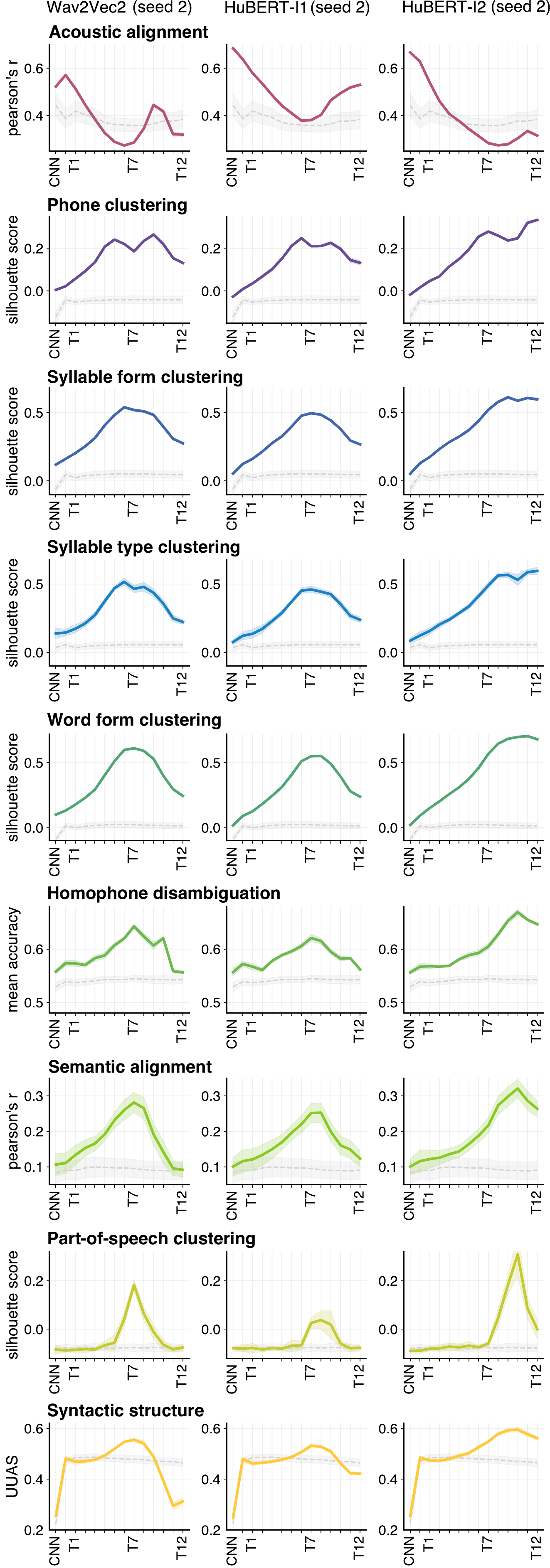}
\vspace{-1em}
\caption{Layerwise scores for all representational probes (rows) and three speech SSL models (columns). Grey dashed lines indicate baseline scores extracted from a Wav2Vec2 model trained on non-speech acoustic scenes. Shading indicates the std. dev. over 5 folds.}
\label{fig:layerwise}
\end{figure}

\noindent this rule, with speech-trained model scores only starting to outperform the nonspeech model around 25k and 50k steps, respectively. 

In the right column of \autoref{fig:learning-w2v2}, we visualize the development of \emph{layerwise} patterns across model training (darker colors indicate higher scores). Here, we mainly observe that layerwise peak patterns are relatively stable across model training, emerging between 10k-50k steps and showing no major shifts throughout further training.

\begin{figure}[H]
    \centering
    \includegraphics[width=\columnwidth]{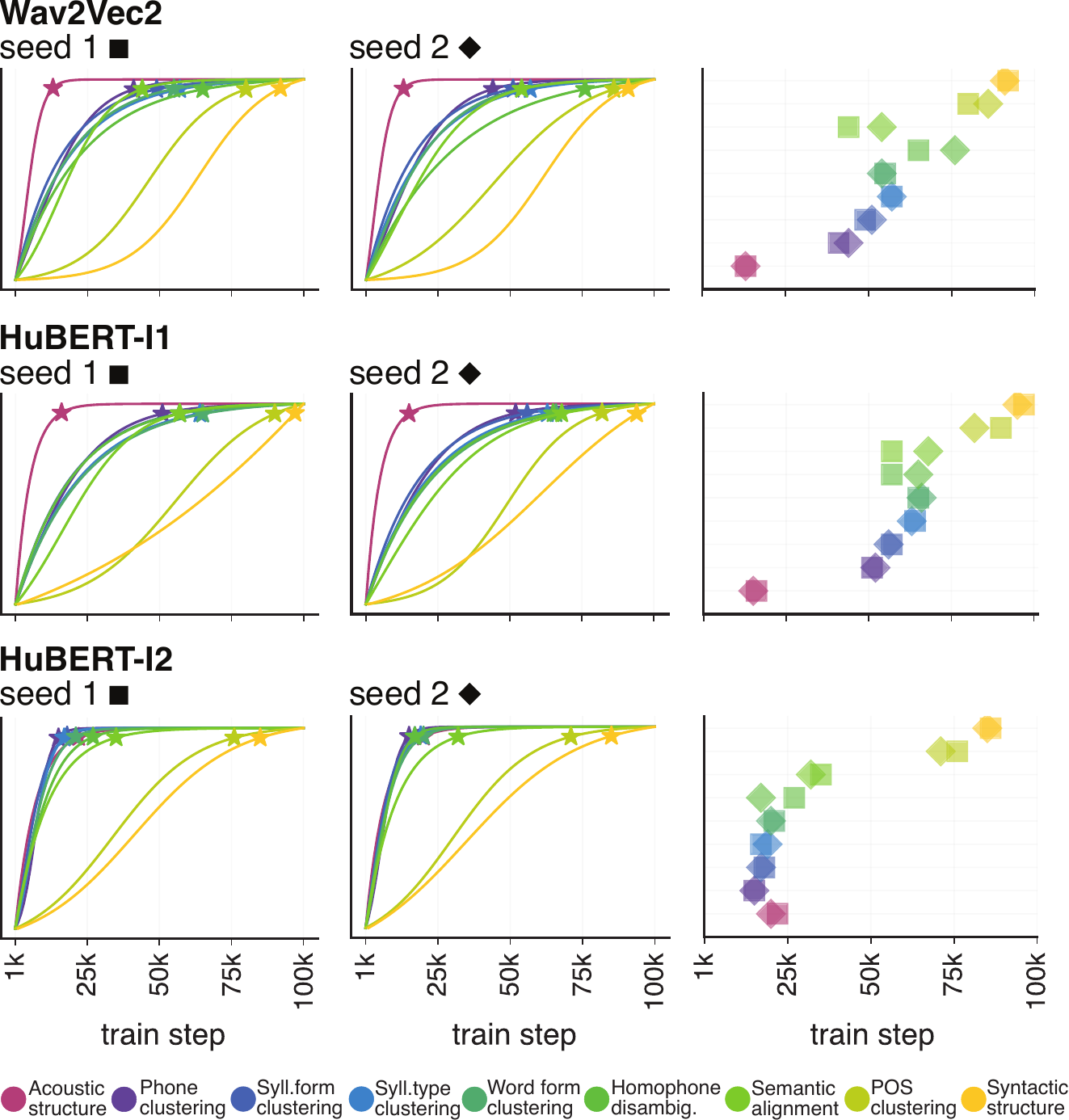}
    \vspace{-1em}
    \caption{Normalized parametric curves\footnoteref{note:sigmoid-fits} visualizing the learning trajectories for all probe scores across training checkpoints, in all six models. Stars indicate the training step where 95\% of the maximum observed score across training is achieved; these steps are again marked in the rightmost plots, with different structure levels on separate rows. Different levels of linguistic structure consistently show distinct learning dynamics across S3M architectures and model seeds, with increased parallelism between levels for HuBERT-I2 models compared to HuBERT-I1 and Wav2Vec2.}
    \label{fig:sigmoids}
\end{figure}

To compare the relative \emph{order} in which linguistic structures at different levels become encoded, we fit parametric curves through the observed probe scores across all analysis checkpoints\footnote{\label{note:sigmoid-fits}We fit the sigmoid function $f(x) = \frac{a}{1 + e^{-k(x-b)}}+c$, using the \texttt{curve\_fit} method from the scipy library \citep{2020SciPy-NMeth} to optimize parameters $a$, $b$, $c$ and $k$.}. Fitted curves for the Wav2Vec2 model in \autoref{fig:learning-w2v2} are visualized as lines through the score datapoints obtained for each checkpoint. Obtaining the learning curves for all probes across models allows us to comprehensively visualize the relative differences between all six models and nine levels of structure (\autoref{fig:sigmoids}). We here observe that the relative differences between learning curves for different probes are generally consistent between different seeds of the same model. The replication of relative learning trajectory patterns between model seeds indicates that distinct levels of structure indeed consistently follow different trajectories. We also observe differences between architectural variations: for Wav2Vec2 and HuBERT-I1, learning trajectories show a similar ordering, with acoustic alignment scores their reaching maximum performance first, followed by a group of phone- syllable- and word-level measures, and finally the part-of-speech and syntactic probes. Conversely, the learning curves of HuBERT-I2 show greater parallelism, diverging from the other models and echoing observed differences in layerwise patterns. The combined pattern of results suggests that HuBERT-I2's divergent behaviour is not a result of its training objective involving the prediction of \emph{categorical} labels (which is shared with HuBERT-I1), but rather of the \emph{iterative refinement} of prediction targets in the HuBERT training process. This complements earlier findings investigating layerwise differences between Wav2Vec2 and HuBERT \citep{huoIterativeRefinementNot2025}, showing that these differences originate in HuBERT-I2's training dynamics. Through

\noindent pseudo-labels, HuBERT-I2 training is guided by highly compressed information from I1 representations. I1 hidden layers abstract away from acoustics and exhibit some degree of linguistic structuring (\autoref{fig:layerwise}), and it is the linguistic structures best encoded in those I1 (layer T6) representations which first reach their maximum performance across I2's training (the phone- and syllable-level scores).

\begin{figure}[H]
    \centering
    \includegraphics[width=\columnwidth]{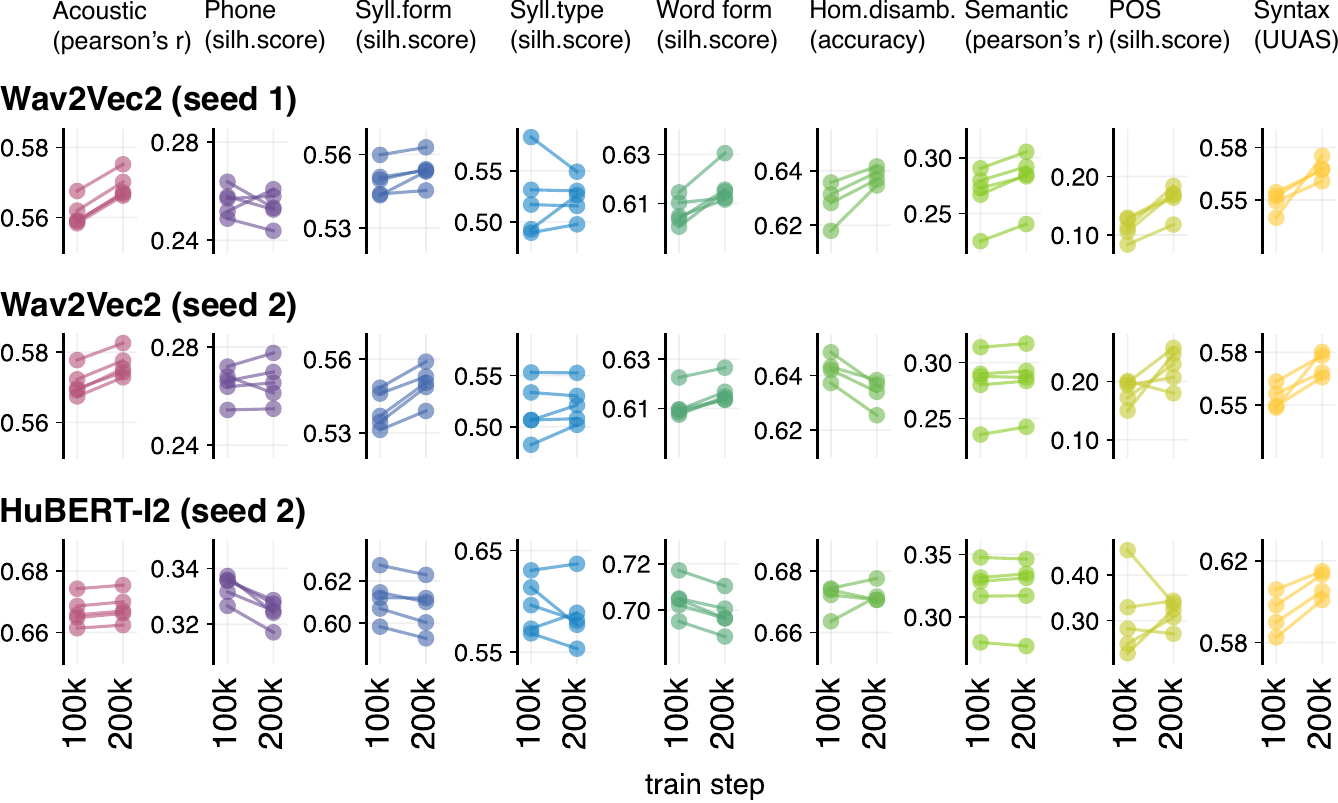}
    \vspace{-1.5em}
    \caption{Relative change in best-layer scores when further training Wav2Vec2 and HuBERT-I2 up to 200K training steps. Note that scores are on different scales across probes; these results are only displayed together to comprehensively visualize relative changes between checkpoints for each probe, not to compare absolute scores between probes. The five lines for each probe connect scores for the same folds of the analysis data.}
    \label{fig:200k-comp}
\end{figure}

Most probe scores across models appear to reach a ceiling before 100K training steps (\autoref{fig:learning-w2v2}, \autoref{fig:sigmoids}). Can additional training further improve the encoding of linguistic structures, or do these scores indeed approximate the maximum obtainable performance for speech representations optimized with SSL objectives? To investigate this, we further trained the Wav2Vec2 and HuBERT-I2 models up to 200K steps, and measure probe performance on the final 200K checkpoint for each. In \autoref{fig:200k-comp} we visualize the results for both Wav2Vec2 and one HuBERT-I2 model\footnote{We here exclude HuBERT-I2 seed 1, because it showed instabilities in training beyond 100K steps; see \autoref{app:model-validation}.}. 

Across models and probes, score improvements are generally very minimal after further training to 200K steps. Only the structural probe, measuring the encoding of syntactic dependencies, shows consistent improvement across models. It is possible that training beyond 100K steps mainly improves the contextualization of model representations beyond the word level, and that other sentence-level information not analyzed here would also show continued improvement (e.g. semantic sentence similarities; \citealp{merkxSemanticSentenceSimilarity2021}). The empirical limits of decoding such information from S3Ms remain to be explored; \citet{orhanEmergencePhonemicSyntactic2026} report that dependency structures extracted from a Wav2Vec2 model (trained on 900 hours of speech up to 400K steps) approximate those from text-based models in accuracy, demonstrating that S3M representations can indeed learn to encode rich syntactic structures after only SSL training.

\section{Discussion \& Conclusions}
Do S3M layerwise representations and learning dynamics reflect the hierarchical organization of linguistic information in speech? We here developed and applied an analysis suite of representational probes to track the encoding of nine levels of linguistic structure across model-internal layers and intermediate training checkpoints.

The observed layerwise patterns and learning trajectories indeed reflect a hierarchy among linguistic structures, though perhaps mostly dominated by timescale: lexical properties related to form, meaning and grammar jointly peak in the same model layers (\autoref{fig:layerwise}), and learning of more abstract word properties (semantics, homophone disambiguation) sometimes precedes word form (\autoref{fig:sigmoids}). Timescale and abstraction levels are often related: while syntactic structures are arguably more abstract, their late development might also reflect that the contextualization of model representations beyond the word timescale only starts later in training, as accurate encoding of both POS and dependency structures requires the integration of some beyond-word context.

Learning dynamics of text-based language models reflect successive stages where token probabilities shift from mimicking unigram to higher n-gram statistics over the course of model training \citep{changWordAcquisitionNeural2022, michaelovLanguageModelBehavioral2025, jumeletBlackBigBoxes2026}. Speech-based models face an additional challenge in first needing to learn what segments of the continuous speech signal in fact constitute the relevant linguistic units to keep track of. Syntactic information consistently emerges last in training (\autoref{fig:sigmoids}); this could indicate that models indeed first need to develop reasonable capability for distinguishing linguistic units (e.g. word forms) before combinatorial structures relating such units can become encoded. However, this hypothesis should ideally be tested in causal experiments (as performed by \citealp{chenSuddenDropsLoss2023a} for syntactic learning in text-based models); such experiments could for example consist of suppressing the encoding of word form across training and observing its effects on the emergence of syntactic structure. The results we report here nevertheless contribute observational evidence that motivates such future work.

The higher-level prediction objective of HuBERT-I2 increases parallelism in learning and also substantially affects its layerwise representations: multiple levels of linguistic structure end up jointly encoded in higher layers (\autoref{fig:layerwise}). Interestingly, the encoding of semantic and syntactic information drops off in HuBERT-I2's final layers, while the encoding of phone, syllable and word forms remains high. Neuroscience research on human speech processing has shown that the brain keeps track of linguistic information across multiple hierarchical levels simultaneously, such that higher-level information can potentially inform lower-level representations \citep{heilbronHierarchyLinguisticPredictions2022, gwilliamsHierarchicalDynamicCoding2025}. Whether the encoding of lower-level linguistic units in HuBERT-I2's final layers similarly benefits from higher-level information represented in earlier layers remains to be explored. From an engineering perspective, the increased linguistic specialization (and degraded acoustic encoding) of representations in models like HuBERT is not necessarily beneficial for all downstream tasks, as noted in recent work developing novel SSL paradigms for the goal of spoken language modeling \citep{poliSpidRLearningFast2025}.

We note some limitations to the analyses presented here. While layerwise similarity to semantic text embeddings generally aligns to other metrics of semantic content \citep{pasadLayerWiseAnalysisSelfSupervised2021a, pasadWhatSelfSupervisedSpeech2024, sauterCuriousCaseVisual2026} and we took measures to avoid analysis confounds, it is possible that other non-semantic (sub)word information encoded by the Fasttext model affects our measurements. Similarly, homophones with identical phonetic transcriptions may nevertheless exhibit acoustic differences in pronunciation \citep{seyfarthAcousticDifferencesMorphologicallydistinct2018}. Hence, the obtained semantic alignment and homophone disambiguation scores might reflect less abstract, form-related aspects of words beyond meaning-related information. Furthermore, although existing work has shown that language-specific pre-training benefits the encoding of linguistic information in the training language \citep{poliModelingInitialState2024, deheerklootsWhatSelfsupervisedSpeech2025, orhanEmergencePhonemicSyntactic2026}, we did not explore whether training on Dutch speech specifically (vs. another language) is necessary for all findings observed here.

More detailed studies of linguistic representation learning in speech SSL models, as well as it causal links to model behavior, could potentially inform the development of speech technology as well as (psycho)linguistic theory. We hope that the analyses we presented here contribute a useful starting point for such investigations. 

\section{Acknowledgements}
This work used the Dutch national e-infrastructure with the support of the SURF Cooperative using grants no. EINF-8324 and EINF-15179. \\
MdHK is funded by the Netherlands Organization for Scientific Research (NWO), through Gravitation Grant 024.001.006 to the Language in Interaction Consortium; MB, HM, CP, GS are funded through NWA-ORC grant NWA.1292.19.399 for ‘InDeep’. We thank three anonymous reviewers for helpful feedback on an earlier version of this manuscript. 

\clearpage

\bibliography{tacl2021}

@article{scikit-learn,
  title={Scikit-learn: Machine Learning in {P}ython},
  author={Pedregosa, F. and Varoquaux, G. and Gramfort, A. and Michel, V.
          and Thirion, Bthrough NWA-ORC grant NWA.1292.19.399 for ‘InDeep’. and Grisel, O. and Blondel, M. and Prettenhofer, P.
          and Weiss, R. and Dubourg, V. and Vanderplas, J. and Passos, A. and
          Cournapeau, D. and Brucher, M. and Perrot, M. and Duchesnay, E.},
  journal={Journal of Machine Learning Research},
  volume={12},
  pages={2825--2830},
  year={2011}
}

@article{keuleersSUBTLEXNLNewMeasure2010,
	title = {{SUBTLEX}-{NL}: {A} new measure for {Dutch} word frequency based on film subtitles},
	volume = {42},
	issn = {1554-3528},
	shorttitle = {{SUBTLEX}-{NL}},
	url = {https://doi.org/10.3758/BRM.42.3.643},
	doi = {10.3758/BRM.42.3.643},
	language = {en},
	number = {3},
	urldate = {2026-04-01},
	journal = {Behavior Research Methods},
	author = {Keuleers, Emmanuel and Brysbaert, Marc and New, Boris},
	month = aug,
	year = {2010},
	pages = {643--650}
}

@inproceedings{mikolov2018advances,
  title={Advances in Pre-Training Distributed Word Representations},
  author={Mikolov, Tomas and Grave, Edouard and Bojanowski, Piotr and Puhrsch, Christian and Joulin, Armand},
  booktitle={Proceedings of the International Conference on Language Resources and Evaluation (LREC 2018)},
  year={2018}
}

@article{seyfarthAcousticDifferencesMorphologicallydistinct2018,
	title = {Acoustic differences in morphologically-distinct homophones},
	volume = {33},
	issn = {2327-3798},
	url = {https://doi.org/10.1080/23273798.2017.1359634},
	doi = {10.1080/23273798.2017.1359634},
	number = {1},
	urldate = {2026-04-01},
	journal = {Language, Cognition and Neuroscience},
	publisher = {Routledge},
	author = {Seyfarth, Scott and Garellek, Marc and Gillingham, Gwendolyn and Ackerman, Farrell and Malouf, Robert},
	year = {2018},
	pages = {32--49}
}

@article{hupkesVisualisationDiagnosticClassifiers2018,
	title = {Visualisation and '{Diagnostic} {Classifiers}' {Reveal} {How} {Recurrent} and {Recursive} {Neural} {Networks} {Process} {Hierarchical} {Structure}},
	volume = {61},
	issn = {1076-9757},
	url = {http://jair.org/index.php/jair/article/view/11196},
	doi = {10.1613/jair.1.11196},
	journal = {Journal of Artificial Intelligence Research},
	author = {Hupkes, Dieuwke and Veldhoen, Sara and Zuidema, Willem},
	month = apr,
	year = {2018},
	pages = {907--926}
}

@inproceedings{deheerklootsWhatSelfsupervisedSpeech2025,
	title = {What do self-supervised speech models know about {Dutch}? {Analyzing} advantages of language-specific pre-training},
	shorttitle = {What do self-supervised speech models know about {Dutch}?},
	url = {https://www.isca-archive.org/interspeech_2025/deheerkloots25_interspeech.html},
	doi = {10.21437/Interspeech.2025-1526},
	urldate = {2026-03-02},
    booktitle = {Proc. Interspeech},
	author = {de Heer Kloots, Marianne and Mohebbi, Hosein and Pouw, Charlotte and Shen, Gaofei and Zuidema, Willem and Bentum, Martijn},
	year = {2025},
	pages = {256--260}
}

@article{kuhlEarlyLanguageAcquisition2004,
	title = {Early language acquisition: cracking the speech code},
	volume = {5},
	copyright = {2004 Springer Nature Limited},
	issn = {1471-0048},
	shorttitle = {Early language acquisition},
	url = {https://www.nature.com/articles/nrn1533},
	doi = {10.1038/nrn1533},
	language = {en},
	number = {11},
	urldate = {2026-03-31},
	journal = {Nature Reviews Neuroscience},
	publisher = {Nature Publishing Group},
	author = {Kuhl, Patricia K.},
	month = nov,
	year = {2004},
	pages = {831--843},
}

@article{werkerINFLUENCESINFANTSPEECH1999,
	title = {Influences on infant speech processing: {Toward} a {New} {Synthesis}},
	shorttitle = {{INFLUENCES} {ON} {INFANT} {SPEECH} {PROCESSING}},
	url = {https://sci-hub.box/10.1146/annurev.psych.50.1.509},
	doi = {10.1146/annurev.psych.50.1.509},
	urldate = {2026-03-31},
	journal = {Annual Review of Psychology},
	author = {Werker, Janet F. and Tees, Richard C.},
	year = {1999}
}

@inproceedings{saphraUnderstandingLearningDynamics2019,
	address = {Minneapolis, Minnesota},
	title = {Understanding {Learning} {Dynamics} {Of} {Language} {Models} with {SVCCA}},
	url = {https://aclanthology.org/N19-1329/},
	doi = {10.18653/v1/N19-1329},
	booktitle = {Proceedings of the 2019 {Conference} of the {North} {American} {Chapter} of the {Association} for {Computational} {Linguistics}: {Human} {Language} {Technologies}, {Volume} 1 ({Long} and {Short} {Papers})},
	publisher = {Association for Computational Linguistics},
	author = {Saphra, Naomi and Lopez, Adam},
	editor = {Burstein, Jill and Doran, Christy and Solorio, Thamar},
	year = {2019},
	pages = {3257--3267}
}

@misc{jumeletBlackBigBoxes2026,
	title = {Black {Big} {Boxes}: {Tracing} {Adjective} {Order} {Preferences} in {Large} {Language} {Models}},
	shorttitle = {Black {Big} {Boxes}},
	url = {http://arxiv.org/abs/2407.02136},
	doi = {10.48550/arXiv.2407.02136},
	urldate = {2026-03-30},
	publisher = {arXiv},
	author = {Jumelet, Jaap and Bylinina, Lisa and Zuidema, Willem and Szymanik, Jakub},
	year = {2026},
	note = {arXiv:2407.02136 [cs]}
}

@inproceedings{michaelovLanguageModelBehavioral2025,
	title = {Language {Model} {Behavioral} {Phases} are {Consistent} {Across} {Architecture}, {Training} {Data}, and {Scale}},
	url = {https://openreview.net/forum?id=HenpVfO3Wp},
	language = {en},
	urldate = {2026-03-30},
	author = {Michaelov, James A. and Levy, Roger P. and Bergen, Ben},
	year = {2025}
}

@inproceedings{merkxSemanticSentenceSimilarity2021,
	title = {Semantic {Sentence} {Similarity}: {Size} does not {Always} {Matter}},
	shorttitle = {Semantic {Sentence} {Similarity}},
	url = {https://www.isca-archive.org/interspeech_2021/merkx21_interspeech.html},
	doi = {10.21437/Interspeech.2021-1464},
	urldate = {2026-03-30},
	author = {Merkx, Danny and Frank, Stefan L. and Ernestus, Mirjam},
	year = {2021},
	pages = {4393--4397}
}

@ARTICLE{2020SciPy-NMeth,
  author  = {Virtanen, Pauli and Gommers, Ralf and Oliphant, Travis E. and
            Haberland, Matt and Reddy, Tyler and Cournapeau, David and
            Burovski, Evgeni and Peterson, Pearu and Weckesser, Warren and
            Bright, Jonathan and {van der Walt}, St{\'e}fan J. and
            Brett, Matthew and Wilson, Joshua and Millman, K. Jarrod and
            Mayorov, Nikolay and Nelson, Andrew R. J. and Jones, Eric and
            Kern, Robert and Larson, Eric and Carey, C J and
            Polat, {\.I}lhan and Feng, Yu and Moore, Eric W. and
            {VanderPlas}, Jake and Laxalde, Denis and Perktold, Josef and
            Cimrman, Robert and Henriksen, Ian and Quintero, E. A. and
            Harris, Charles R. and Archibald, Anne M. and
            Ribeiro, Ant{\^o}nio H. and Pedregosa, Fabian and
            {van Mulbregt}, Paul and {SciPy 1.0 Contributors}},
  title   = {{{SciPy} 1.0: Fundamental Algorithms for Scientific
            Computing in Python}},
  journal = {Nature Methods},
  year    = {2020},
  volume  = {17},
  pages   = {261--272},
  adsurl  = {https://rdcu.be/b08Wh},
  doi     = {10.1038/s41592-019-0686-2},
}

@inproceedings{bentumWordStressSelfsupervised2025,
	title = {Word stress in self-supervised speech models: {A} cross-linguistic comparison},
	shorttitle = {Word stress in self-supervised speech models},
	url = {https://www.isca-archive.org/interspeech_2025/bentum25_interspeech.html},
	doi = {10.21437/Interspeech.2025-106},
	urldate = {2026-03-30},
	author = {Bentum, Martijn and ten Bosch, Louis and Lentz, Tomas O.},
	year = {2025},
	pages = {251--255}
}

@misc{wolfHuggingFacesTransformersStateoftheart2020,
	title = {{HuggingFace}'s {Transformers}: {State}-of-the-art {Natural} {Language} {Processing}},
	shorttitle = {{HuggingFace}'s {Transformers}},
	url = {http://arxiv.org/abs/1910.03771},
	doi = {10.48550/arXiv.1910.03771},
	urldate = {2025-02-19},
	publisher = {arXiv},
	author = {Wolf, Thomas and Debut, Lysandre and Sanh, Victor and Chaumond, Julien and Delangue, Clement and Moi, Anthony and Cistac, Pierric and Rault, Tim and Louf, Rémi and Funtowicz, Morgan and Davison, Joe and Shleifer, Sam and Platen, Patrick von and Ma, Clara and Jernite, Yacine and Plu, Julien and Xu, Canwen and Scao, Teven Le and Gugger, Sylvain and Drame, Mariama and Lhoest, Quentin and Rush, Alexander M.},
	year = {2020}
}

@inproceedings{mcfeeLibrosaAudioMusic2015,
	address = {Austin, Texas},
	title = {librosa: {Audio} and {Music} {Signal} {Analysis} in {Python}},
	shorttitle = {librosa},
	url = {https://doi.curvenote.com/10.25080/Majora-7b98e3ed-003},
	doi = {10.25080/Majora-7b98e3ed-003},
	language = {en},
	author = {McFee, Brian and Raffel, Colin and Liang, Dawen and Ellis, Daniel and McVicar, Matt and Battenberg, Eric and Nieto, Oriol},
	year = {2015},
	pages = {18--24}
}

@inproceedings{delafuenteLayerwiseAnalysisMandarin2024a,
	title = {A layer-wise analysis of {Mandarin} and {English} suprasegmentals in {SSL} speech models},
	url = {https://www.isca-archive.org/interspeech_2024/delafuente24_interspeech.html},
	doi = {10.21437/Interspeech.2024-2341},
	language = {en},
	urldate = {2025-08-21},
	booktitle = {Interspeech 2024},
	publisher = {ISCA},
	author = {De La Fuente, Anton and Jurafsky, Dan},
	month = sep,
	year = {2024},
	pages = {1290--1294}
}

@inproceedings{choiLeveragingAllophonySelfSupervised2025,
	address = {Albuquerque, New Mexico},
	title = {Leveraging {Allophony} in {Self}-{Supervised} {Speech} {Models} for {Atypical} {Pronunciation} {Assessment}},
	isbn = {979-8-89176-189-6},
	url = {https://aclanthology.org/2025.naacl-long.132/},
	doi = {10.18653/v1/2025.naacl-long.132},
	urldate = {2026-03-29},
	booktitle = {Proceedings of the 2025 {Conference} of the {Nations} of the {Americas} {Chapter} of the {Association} for {Computational} {Linguistics}: {Human} {Language} {Technologies} ({Volume} 1: {Long} {Papers})},
	publisher = {Association for Computational Linguistics},
	author = {Choi, Kwanghee and Yeo, Eunjung and Chang, Kalvin and Watanabe, Shinji and Mortensen, David R},
	editor = {Chiruzzo, Luis and Ritter, Alan and Wang, Lu},
	year = {2025},
	pages = {2613--2628}
}

@inproceedings{martinProbingSelfsupervisedSpeech2023,
	title = {Probing {Self}-supervised {Speech} {Models} for {Phonetic} and {Phonemic} {Information}: {A} {Case} {Study} in {Aspiration}},
	shorttitle = {Probing {Self}-supervised {Speech} {Models} for {Phonetic} and {Phonemic} {Information}},
	url = {https://www.isca-archive.org/interspeech_2023/martin23_interspeech.html},
	doi = {10.21437/Interspeech.2023-2359},
	language = {en},
	urldate = {2024-03-11},
	booktitle = {{INTERSPEECH} 2023},
	publisher = {ISCA},
	author = {Martin, Kinan and Gauthier, Jon and Breiss, Canaan and Levy, Roger},
	month = aug,
	year = {2023},
	pages = {251--255}
}

@inproceedings{mohamedOrthogonalityIsotropySpeaker2024,
	title = {Orthogonality and isotropy of speaker and phonetic information in self-supervised speech representations},
	url = {https://www.isca-archive.org/interspeech_2024/mohamed24_interspeech.html},
	doi = {10.21437/Interspeech.2024-1054},
	urldate = {2026-03-29},
	author = {Mohamed, Mukhtar and Liu, Oli Danyi and Tang, Hao and Goldwater, Sharon},
	year = {2024},
	pages = {3625--3629}
}

@article{feldmanRoleDevelopingLexicon2013,
	address = {US},
	title = {A role for the developing lexicon in phonetic category acquisition},
	volume = {120},
	issn = {1939-1471},
	doi = {10.1037/a0034245},
	number = {4},
	journal = {Psychological Review},
	publisher = {American Psychological Association},
	author = {Feldman, Naomi H. and Griffiths, Thomas L. and Goldwater, Sharon and Morgan, James L.},
	year = {2013},
	pages = {751--778}
}

@article{gleitmanStructuralSourcesVerb1990,
	title = {The {Structural} {Sources} of {Verb} {Meanings}},
	volume = {1},
	issn = {1048-9223},
	url = {https://doi.org/10.1207/s15327817la0101_2},
	doi = {10.1207/s15327817la0101_2},
	number = {1},
	urldate = {2026-03-29},
	journal = {Language Acquisition},
	publisher = {Routledge},
	author = {Gleitman, Lila},
	month = jan,
	year = {1990},
	note = {\_eprint: https://doi.org/10.1207/s15327817la0101\_2},
	pages = {3--55},
}

@book{perkellInvarianceVariabilitySpeech1986,
	title = {Invariance and {Variability} in {Speech} {Processes}},
	isbn = {978-1-317-76829-6},
	language = {en},
	publisher = {Psychology Press},
	author = {Perkell, J. S. and Klatt, D. H.},
	year = {1986}
}

@incollection{UnderstandingSpeechContext2016,
	title = {Understanding {Speech} in the {Context} of {Variability}},
    author = {Heald, Shannon and Klos, Serena and Nusbaum, Howard},
	url = {https://www.sciencedirect.com/science/chapter/edited-volume/abs/pii/B9780124077942000171},
	doi = {10.1016/B978-0-12-407794-2.00017-1},
	language = {en-US},
	urldate = {2026-03-29},
	booktitle = {Neurobiology of {Language}},
	publisher = {Academic Press},
	year = {2016},
	doi = {10.1016/B978-0-12-407794-2.00017-1},
	pages = {195--208}
}

@article{schatzEarlyPhoneticLearning2021,
	title = {Early phonetic learning without phonetic categories: {Insights} from large-scale simulations on realistic input},
	volume = {118},
	shorttitle = {Early phonetic learning without phonetic categories},
	url = {https://www.pnas.org/doi/full/10.1073/pnas.2001844118},
	doi = {10.1073/pnas.2001844118},
	number = {7},
	urldate = {2026-03-29},
	journal = {Proceedings of the National Academy of Sciences},
	publisher = {Proceedings of the National Academy of Sciences},
	author = {Schatz, Thomas and Feldman, Naomi H. and Goldwater, Sharon and Cao, Xuan-Nga and Dupoux, Emmanuel},
	year = {2021},
	pages = {e2001844118}
}

@article{mohamedSelfSupervisedSpeechRepresentation2022,
	title = {Self-{Supervised} {Speech} {Representation} {Learning}: {A} {Review}},
	volume = {16},
	issn = {1941-0484},
	shorttitle = {Self-{Supervised} {Speech} {Representation} {Learning}},
	doi = {10.1109/JSTSP.2022.3207050},
	number = {6},
	journal = {IEEE Journal of Selected Topics in Signal Processing},
	author = {Mohamed, Abdelrahman and Lee, Hung-yi and Borgholt, Lasse and Havtorn, Jakob D. and Edin, Joakim and Igel, Christian and Kirchhoff, Katrin and Li, Shang-Wen and Livescu, Karen and Maaløe, Lars and Sainath, Tara N. and Watanabe, Shinji},
	year = {2022},
	note = {Conference Name: IEEE Journal of Selected Topics in Signal Processing},
	pages = {1179--1210}
}

@misc{teamOmnilingualASROpenSource2025a,
	title = {Omnilingual {ASR}: {Open}-{Source} {Multilingual} {Speech} {Recognition} for 1600+ {Languages}},
	shorttitle = {Omnilingual {ASR}},
	url = {https://arxiv.org/abs/2511.09690v1},
	author = {{Omnilingual ASR Team} and Keren, Gil and Kozhevnikov, Artyom and Meng, Yen and Ropers, Christophe and Setzler, Matthew and Wang, Skyler and Adebara, Ife and Auli, Michael and Balioglu, Can and Chan, Kevin and Cheng, Chierh and Chuang, Joe and Droof, Caley and Duppenthaler, Mark and Duquenne, Paul-Ambroise and Erben, Alexander and Gao, Cynthia and Gonzalez, Gabriel Mejia and Lyu, Kehan and Miglani, Sagar and Pratap, Vineel and Sadagopan, Kaushik Ram and Saleem, Safiyyah and Turkatenko, Arina and Ventayol-Boada, Albert and Yong, Zheng-Xin and Chung, Yu-An and Maillard, Jean and Moritz, Rashel and Mourachko, Alexandre and Williamson, Mary and Yates, Shireen},
	year = {2025}
}

@article{aroraLandscapeSpokenLanguage2025,
	title = {On {The} {Landscape} of {Spoken} {Language} {Models}: {A} {Comprehensive} {Survey}},
	issn = {2835-8856},
	shorttitle = {On {The} {Landscape} of {Spoken} {Language} {Models}},
	url = {https://openreview.net/forum?id=BvxaP3sVbA},
	language = {en},
	journal = {Transactions on Machine Learning Research},
	author = {Arora, Siddhant and Chang, Kai-Wei and Chien, Chung-Ming and Peng, Yifan and Wu, Haibin and Adi, Yossi and Dupoux, Emmanuel and Lee, Hung-yi and Livescu, Karen and Watanabe, Shinji},
	year = {2025}
}

@article{chenWavLMLargeScaleSelfSupervised2022a,
	title = {{WavLM}: {Large}-{Scale} {Self}-{Supervised} {Pre}-{Training} for {Full} {Stack} {Speech} {Processing}},
	volume = {16},
	issn = {1941-0484},
	shorttitle = {{WavLM}},
	url = {https://ieeexplore.ieee.org/abstract/document/9814838},
	doi = {10.1109/JSTSP.2022.3188113},
	number = {6},
	urldate = {2026-03-29},
	journal = {IEEE Journal of Selected Topics in Signal Processing},
	author = {Chen, Sanyuan and Wang, Chengyi and Chen, Zhengyang and Wu, Yu and Liu, Shujie and Chen, Zhuo and Li, Jinyu and Kanda, Naoyuki and Yoshioka, Takuya and Xiao, Xiong and Wu, Jian and Zhou, Long and Ren, Shuo and Qian, Yanmin and Qian, Yao and Wu, Jian and Zeng, Michael and Yu, Xiangzhan and Wei, Furu},
	year = {2022},
	pages = {1505--1518}
}

@article{poliSpidRLearningFast2025,
	title = {{SpidR}: {Learning} {Fast} and {Stable} {Linguistic} {Units} for {Spoken} {Language} {Models} {Without} {Supervision}},
	issn = {2835-8856},
	shorttitle = {{SpidR}},
	url = {https://openreview.net/forum?id=E7XAFBpfZs},
	language = {en},
	urldate = {2026-03-29},
	journal = {Transactions on Machine Learning Research},
	author = {Poli, Maxime and Luthra, Mahi and Benchekroun, Youssef and Higuchi, Yosuke and Gleize, Martin and Shen, Jiayi and Algayres, Robin and Chung, Yu-An and Assran, Mido and Pino, Juan and Dupoux, Emmanuel},
	year = {2025}
}

@article{zuidemaSyllableFrequencyList,
	title = {A syllable frequency list for {Dutch}},
	language = {en},
	author = {Zuidema, Willem},
        year = 2009,
        journal = {ILLC Preprint Series (PP-2009-50)},
        url = {https://staff.fnwi.uva.nl/w.zuidema/papers/zuidema09tr-syllfreq.pdf}
}

@inproceedings{vaessenSelfsupervisedLearningSpeech2025,
	title = {Self-supervised learning of speech representations with {Dutch} archival data},
	url = {https://www.isca-archive.org/interspeech_2025/vaessen25_interspeech.html},
	doi = {10.21437/Interspeech.2025-463},
	urldate = {2026-03-26},
	author = {Vaessen, Nik and Ordelman, Roeland and van Leeuwen, David A.},
	year = {2025},
	pages = {1208--1212}
}

@inproceedings{bransMultiSimLex2026,
	title = {Multi-{SimLex} for {Dutch}: {Benchmarking} {Embedding}- and {Prompt}-{Based} {Model} {Performance} on {Semantic} {Similarity}},
	booktitle = {LREC 2026},
    year = {2026},
	publisher = {ELRA},
	author = {Brans, Lizzy and Bloem, Jelke}
}

@inproceedings{vansonIFADVCorpusFree2008,
	title = {The {IFADV} {Corpus}: a {Free} {Dialog} {Video} {Corpus}},
	shorttitle = {The {IFADV} {Corpus}},
	booktitle = {LREC 2008},
    year = {2008},
	publisher = {ELRA},
	author = {van Son, Rob and Wesseling, Wieneke and Sanders, Eric and van den Heuvel, Henk},
	editor = {Calzolari, Nicoletta and Choukri, Khalid and Maegaard, Bente and Mariani, Joseph and Odijk, Jan and Piperidis, Stelios and Tapias, Daniel}
}

@misc{rollCategorizeEarlyIntegrate2026,
	title = {Categorize {Early}, {Integrate} {Late}: {Divergent} {Processing} {Strategies} in {Automatic} {Speech} {Recognition}},
	shorttitle = {Categorize {Early}, {Integrate} {Late}},
	url = {http://arxiv.org/abs/2601.06972},
	doi = {10.48550/arXiv.2601.06972},
	urldate = {2026-03-20},
	publisher = {arXiv},
	author = {Roll, Nathan and Bhalerao, Pranav and Bartelds, Martijn and Pawar, Arjun and Tatsumi, Yuka and Ogunremi, Tolulope and Shani, Chen and Graham, Calbert and Sumner, Meghan and Jurafsky, Dan},
	year = {2026}
}

@inproceedings{heLayerwiseMinimalPair2025,
	address = {Suzhou, China},
	title = {Layer-wise {Minimal} {Pair} {Probing} {Reveals} {Contextual} {Grammatical}-{Conceptual} {Hierarchy} in {Speech} {Representations}},
	isbn = {979-8-89176-332-6},
	url = {https://aclanthology.org/2025.emnlp-main.1790/},
	doi = {10.18653/v1/2025.emnlp-main.1790},
	booktitle = {Proceedings of the 2025 {Conference} on {Empirical} {Methods} in {Natural} {Language} {Processing}},
	publisher = {Association for Computational Linguistics},
	author = {He, Linyang and Wang, Qiaolin and Jiang, Xilin and Mesgarani, Nima},
	editor = {Christodoulopoulos, Christos and Chakraborty, Tanmoy and Rose, Carolyn and Peng, Violet},
	month = nov,
	year = {2025},
	pages = {35338--35353}
}

@inproceedings{gauthierEmergentMorphophonologicalRepresentations2025,
	address = {Suzhou, China},
	title = {Emergent morpho-phonological representations in self-supervised speech models},
	isbn = {979-8-89176-332-6},
	url = {https://aclanthology.org/2025.emnlp-main.1425/},
	doi = {10.18653/v1/2025.emnlp-main.1425},
	urldate = {2026-03-20},
	booktitle = {Proceedings of the 2025 {Conference} on {Empirical} {Methods} in {Natural} {Language} {Processing}},
	publisher = {Association for Computational Linguistics},
	author = {Gauthier, Jon and Breiss, Canaan and Leonard, Matthew K and Chang, Edward F.},
	editor = {Christodoulopoulos, Christos and Chakraborty, Tanmoy and Rose, Carolyn and Peng, Violet},
	year = {2025},
	pages = {28067--28086}
}

@misc{orhanEmergencePhonemicSyntactic2026,
	title = {Emergence of {Phonemic}, {Syntactic}, and {Semantic} {Representations} in {Artificial} {Neural} {Networks}},
	url = {http://arxiv.org/abs/2601.18617},
	doi = {10.48550/arXiv.2601.18617},
	publisher = {arXiv},
	author = {Orhan, Pierre and Diego-Simón, Pablo and Chemla, Emmnanuel and Lakretz, Yair and Boubenec, Yves and King, Jean-Rémi},
	year = {2026}
}

@inproceedings{sauterCuriousCaseVisual2026,
	title = {The {Curious} {Case} of {Visual} {Grounding}: {Different} {Effects} for {Speech}- and {Text}-based {Language} {Encoders}},
	shorttitle = {The {Curious} {Case} of {Visual} {Grounding}},
	url = {http://arxiv.org/abs/2509.15837},
	doi = {10.48550/arXiv.2509.15837},
    booktitle = {{ICASSP} 2026 - 2026 {IEEE} {International} {Conference} on {Acoustics}, {Speech} and {Signal} {Processing} ({ICASSP})},
	author = {Sauter, Adrian and Zuidema, Willem and Kloots, Marianne de Heer},
	year = {2026},
}

@article{tharwatLinearDiscriminantAnalysis2017,
	title = {Linear discriminant analysis: {A} detailed tutorial},
	volume = {30},
	issn = {0921-7126},
	shorttitle = {Linear discriminant analysis},
	url = {https://journals.sagepub.com/action/showAbstract},
	doi = {10.3233/AIC-170729},
	language = {EN},
	number = {2},
	urldate = {2026-03-02},
	journal = {AI Communications},
	publisher = {SAGE Publications},
	author = {Tharwat, Alaa and Gaber, Tarek and Ibrahim, Abdelhameed and Hassanien, Aboul Ella},
	month = may,
	year = {2017},
	pages = {169--190},
}

@inproceedings{choiSelfSupervisedSpeechRepresentations2024a,
	title = {Self-{Supervised} {Speech} {Representations} are {More} {Phonetic} than {Semantic}},
	url = {https://www.isca-archive.org/interspeech_2024/choi24b_interspeech.html},
	doi = {10.21437/Interspeech.2024-1157},
	urldate = {2025-04-08},
	author = {Choi, Kwanghee and Pasad, Ankita and Nakamura, Tomohiko and Fukayama, Satoru and Livescu, Karen and Watanabe, Shinji},
	year = {2024},
	pages = {4578--4582}
}

@inproceedings{huoIterativeRefinementNot2025,
	title = {Iterative {Refinement}, {Not} {Training} {Objective}, {Makes} {HuBERT} {Behave} {Differently} from wav2vec 2.0},
	url = {https://www.isca-archive.org/interspeech_2025/huo25_interspeech.html},
	doi = {10.21437/Interspeech.2025-514},
	urldate = {2026-03-20},
	author = {Huo, Robin and Dunbar, Ewan},
	year = {2025},
	pages = {261--265}
}

@article{algayresDPParseFindingWord2022,
	title = {{DP}-{Parse}: {Finding} {Word} {Boundaries} from {Raw} {Speech} with an {Instance} {Lexicon}},
	volume = {10},
	issn = {2307-387X},
	shorttitle = {{DP}-{Parse}},
	url = {https://doi.org/10.1162/tacl_a_00505},
	doi = {10.1162/tacl_a_00505},
	urldate = {2025-02-18},
	journal = {Transactions of the Association for Computational Linguistics},
	author = {Algayres, Robin and Ricoul, Tristan and Karadayi, Julien and Laurençon, Hugo and Zaiem, Salah and Mohamed, Abdelrahman and Sagot, Benoît and Dupoux, Emmanuel},
	month = sep,
	year = {2022},
	pages = {1051--1065}
}

@article{poliModelingInitialState2024,
	title = {Modeling the initial state of early phonetic learning in infants},
	volume = {5},
	issn = {2771-7976},
	url = {http://ldr.lps.library.cmu.edu/article/id/717/},
	doi = {10.34842/y89t-6q31},
	language = {eng},
	number = {1},
	urldate = {2025-02-19},
	journal = {Language Development Research},
	author = {Poli, Maxime and Schatz, Thomas and Dupoux, Emmanuel and Lavechin, Marvin},
	month = aug,
	year = {2024}
}

@article{lavechinSimulatingEarlyPhonetic2025,
	title = {Simulating {Early} {Phonetic} and {Word} {Learning} {Without} {Linguistic} {Categories}},
	volume = {28},
	issn = {1363-755X},
	url = {https://onlinelibrary.wiley.com/doi/abs/10.1111/desc.13606},
	doi = {10.1111/desc.13606},
	number = {2},
	urldate = {2025-02-11},
	journal = {Developmental Science},
	author = {Lavechin, Marvin and de Seyssel, Maureen and Titeux, Hadrien and Wisniewski, Guillaume and Bredin, Hervé and Cristia, Alejandrina and Dupoux, Emmanuel},
	month = mar,
	year = {2025}
}

@inproceedings{chenSuddenDropsLoss2023a,
	title = {Sudden {Drops} in the {Loss}: {Syntax} {Acquisition}, {Phase} {Transitions}, and {Simplicity} {Bias} in {MLMs}},
	shorttitle = {Sudden {Drops} in the {Loss}},
	url = {https://openreview.net/forum?id=MO5PiKHELW},
	author = {Chen, Angelica and Shwartz-Ziv, Ravid and Cho, Kyunghyun and Leavitt, Matthew L. and Saphra, Naomi},
        booktitle = {ICLR},
	month = oct,
	year = {2024}
}

@article{mcclellandTRACEModelSpeech1986a,
	title = {The {TRACE} model of speech perception},
	volume = {18},
	issn = {0010-0285},
	url = {https://www.sciencedirect.com/science/article/pii/0010028586900150},
	doi = {10.1016/0010-0285(86)90015-0},
	number = {1},
	urldate = {2025-03-14},
	journal = {Cognitive Psychology},
	author = {McClelland, James L and Elman, Jeffrey L},
	month = jan,
	year = {1986},
	pages = {1--86},
}

@book{rogersSemanticCognitionParallel2004,
	title = {Semantic {Cognition}: {A} {Parallel} {Distributed} {Processing} {Approach}},
	isbn = {978-0-262-18239-3},
	shorttitle = {Semantic {Cognition}},
	language = {en},
	publisher = {MIT Press},
	author = {Rogers, Timothy T. and McClelland, James L.},
	year = {2004}
}

@inproceedings{pasadComparativeLayerWiseAnalysis2023,
	title = {Comparative {Layer}-{Wise} {Analysis} of {Self}-{Supervised} {Speech} {Models}},
	url = {https://ieeexplore.ieee.org/abstract/document/10096149?casa_token=bDtLdXJ3q-4AAAAA:HzWpZAgcqZAZWy87NtLAIuQBWjNUkXGovsp1aokTaYsWzPTxSbHBaPL9T32YxPvQRPXavpowU5s},
	doi = {10.1109/ICASSP49357.2023.10096149},
	booktitle = {{ICASSP} 2023 - 2023 {IEEE} {International} {Conference} on {Acoustics}, {Speech} and {Signal} {Processing} ({ICASSP})},
	author = {Pasad, Ankita and Shi, Bowen and Livescu, Karen},
	month = jun,
	year = {2023},
	note = {ISSN: 2379-190X},
	pages = {1--5}
}

@inproceedings{pasadLayerWiseAnalysisSelfSupervised2021a,
	title = {Layer-{Wise} {Analysis} of a {Self}-{Supervised} {Speech} {Representation} {Model}},
	doi = {10.1109/ASRU51503.2021.9688093},
	urldate = {2025-03-14},
	booktitle = {2021 {IEEE} {Automatic} {Speech} {Recognition} and {Understanding} {Workshop} ({ASRU})},
	author = {Pasad, Ankita and Chou, Ju-Chieh and Livescu, Karen},
	month = dec,
	year = {2021},
	pages = {914--921},
}

@article{pasadWhatSelfSupervisedSpeech2024,
	title = {What {Do} {Self}-{Supervised} {Speech} {Models} {Know} {About} {Words}?},
	volume = {12},
	issn = {2307-387X},
	doi = {10.1162/tacl_a_00656},
	urldate = {2024-10-10},
	journal = {TACL},
	author = {Pasad, Ankita and Chien, Chung-Ming and Settle, Shane and Livescu, Karen},
	year = {2024},
	pages = {372--391}
}

@inproceedings{deheerklootsHumanlikeLinguisticBiases2024,
	title = {Human-like {Linguistic} {Biases} in {Neural} {Speech} {Models}: {Phonetic} {Categorization} and {Phonotactic} {Constraints} in {Wav2Vec2}.0},
	shorttitle = {Human-like {Linguistic} {Biases} in {Neural} {Speech} {Models}},
	doi = {10.21437/Interspeech.2024-2490},
        year = 2024,
        booktitle = {Proc. Interspeech},
	urldate = {2024-10-10},
	author = {de Heer Kloots, Marianne and Zuidema, Willem},
	pages = {4593--4597},
}

@article{hsuHuBERTSelfSupervisedSpeech2021,
	title = {{HuBERT}: {Self}-{Supervised} {Speech} {Representation} {Learning} by {Masked} {Prediction} of {Hidden} {Units}},
	volume = {29},
	issn = {2329-9290},
	shorttitle = {{HuBERT}},
	url = {https://doi.org/10.1109/TASLP.2021.3122291},
	doi = {10.1109/TASLP.2021.3122291},
	journal = {IEEE/ACM Trans. Audio, Speech and Lang. Proc.},
	author = {Hsu, Wei-Ning and Bolte, Benjamin and Tsai, Yao-Hung Hubert and Lakhotia, Kushal and Salakhutdinov, Ruslan and Mohamed, Abdelrahman},
	month = oct,
	year = {2021},
	pages = {3451--3460}
}

@article{dupouxCognitiveScienceEra2018,
	title = {Cognitive science in the era of artificial intelligence: {A} roadmap for reverse-engineering the infant language-learner},
	volume = {173},
	issn = {00100277},
	shorttitle = {Cognitive science in the era of artificial intelligence},
	url = {https://linkinghub.elsevier.com/retrieve/pii/S0010027717303013},
	doi = {10.1016/j.cognition.2017.11.008},
	language = {en},
	urldate = {2022-12-12},
	journal = {Cognition},
	author = {Dupoux, Emmanuel},
	month = apr,
	year = {2018},
	pages = {43--59}
}

@article{arnonStartingBigApproach2021,
	title = {The {Starting} {Big} approach to language learning},
	volume = {48},
	copyright = {Copyright © The Author(s), 2021. Published by Cambridge University Press},
	issn = {03050009},
	url = {https://www.proquest.com/docview/2572817524/abstract/677D21D2ACCD493DPQ/1},
	doi = {10.1017/S0305000921000386},
	language = {English},
	number = {5},
	urldate = {2025-03-14},
	journal = {Journal of Child Language},
	author = {Arnon, Inbal},
	month = sep,
	year = {2021},
	note = {Num Pages: 937-958
Place: Cambridge, United Kingdom
Publisher: Cambridge University Press
Section: Article},
	pages = {937--958},
}

@book{tomaselloConstructingLanguageUsageBased2003,
	title = {Constructing a {Language}: {A} {Usage}-{Based} {Theory} of {Language} {Acquisition}},
	isbn = {978-0-674-01764-1},
	shorttitle = {Constructing a {Language}},
	language = {en},
	publisher = {Harvard University Press},
	author = {Tomasello, Michael},
	month = mar,
	year = {2003}
}

@article{werkerPerceptualBeginningsLanguage2018,
	title = {Perceptual beginnings to language acquisition},
	volume = {39},
	issn = {0142-7164, 1469-1817},
	url = {https://www.cambridge.org/core/journals/applied-psycholinguistics/article/abs/perceptual-beginnings-to-language-acquisition/DE2FF7F5716CA76834F5051D6C219D69},
	doi = {10.1017/S0142716418000152},
	language = {en},
	number = {4},
	urldate = {2025-03-14},
	journal = {Applied Psycholinguistics},
	author = {Werker, Janet F.},
	month = jul,
	year = {2018},
	pages = {703--728},
}

@article{kuhlNewViewLanguage2000,
	title = {A new view of language acquisition},
	volume = {97},
	url = {https://www.pnas.org/doi/abs/10.1073/pnas.97.22.11850},
	doi = {10.1073/pnas.97.22.11850},
	number = {22},
	urldate = {2025-03-14},
	journal = {Proceedings of the National Academy of Sciences},
	author = {Kuhl, Patricia K.},
	month = oct,
	year = {2000},
	note = {Publisher: Proceedings of the National Academy of Sciences},
	pages = {11850--11857}
}

@misc{gwilliamsHierarchicalDynamicCoding2025,
	title = {Hierarchical dynamic coding coordinates speech comprehension in the human brain},
	url = {https://www.biorxiv.org/content/10.1101/2024.04.19.590280v2},
	doi = {10.1101/2024.04.19.590280},
	language = {en},
	urldate = {2025-03-14},
	publisher = {bioRxiv},
	author = {Gwilliams, Laura and Marantz, Alec and Poeppel, David and King, Jean-Remi},
	month = mar,
	year = {2025}
}

@article{heilbronHierarchyLinguisticPredictions2022,
	title = {A hierarchy of linguistic predictions during natural language comprehension},
	volume = {119},
	url = {https://www.pnas.org/doi/abs/10.1073/pnas.2201968119},
	doi = {10.1073/pnas.2201968119},
	number = {32},
	urldate = {2025-03-14},
	journal = {Proceedings of the National Academy of Sciences},
	author = {Heilbron, Micha and Armeni, Kristijan and Schoffelen, Jan-Mathijs and Hagoort, Peter and de Lange, Floris P.},
	month = aug,
	year = {2022},
	note = {Publisher: Proceedings of the National Academy of Sciences},
	pages = {e2201968119}
}

@article{marslen-wilsonSentencePerceptionInteractive1975,
	title = {Sentence {Perception} as an {Interactive} {Parallel} {Process}},
	copyright = {1975 by the American Association for the Advancement of Science},
	language = {EN},
	urldate = {2025-03-14},
	journal = {Science},
	author = {Marslen-Wilson, William D.},
	month = jul,
	year = {1975},
        url = {https://doi.org/10.1126/science.189.4198.226},
	note = {Publisher: American Association for the Advancement of Science}
}

@article{elmanCognitivePenetrationMechanisms1988,
	title = {Cognitive penetration of the mechanisms of perception: {Compensation} for coarticulation of lexically restored phonemes},
	volume = {27},
	issn = {0749-596X},
	shorttitle = {Cognitive penetration of the mechanisms of perception},
	url = {https://www.sciencedirect.com/science/article/pii/0749596X8890071X},
	doi = {10.1016/0749-596X(88)90071-X},
	number = {2},
	urldate = {2025-03-14},
	journal = {Journal of Memory and Language},
	author = {Elman, Jeffrey L and McClelland, James L},
	month = apr,
	year = {1988},
	pages = {143--165}
}

@article{friedericiNeuralBasisAuditory2002,
	title = {Towards a neural basis of auditory sentence processing},
	volume = {6},
	issn = {1364-6613, 1879-307X},
	url = {https://www.cell.com/trends/cognitive-sciences/abstract/S1364-6613(00)01839-8},
	doi = {10.1016/S1364-6613(00)01839-8},
	language = {English},
	number = {2},
	urldate = {2025-03-14},
	journal = {Trends in Cognitive Sciences},
	author = {Friederici, Angela D.},
	month = feb,
	year = {2002},
	pmid = {15866191},
	note = {Publisher: Elsevier},
	pages = {78--84}
}

@article{weberModelsSpokenwordRecognition2012,
	title = {Models of spoken‐word recognition},
	volume = {3},
	copyright = {http://onlinelibrary.wiley.com/termsAndConditions\#vor},
	issn = {1939-5078, 1939-5086},
	url = {https://wires.onlinelibrary.wiley.com/doi/10.1002/wcs.1178},
	doi = {10.1002/wcs.1178},
	language = {en},
	number = {3},
	urldate = {2025-03-14},
	journal = {WIREs Cognitive Science},
	author = {Weber, Andrea and Scharenborg, Odette},
	month = may,
	year = {2012},
	pages = {387--401}
}

@article{ardilaCommonVoiceMassivelyMultilingual2020,
  title = {Common {{Voice}}: {{A Massively-Multilingual Speech Corpus}}},
  shorttitle = {Common {{Voice}}},
  author = {Ardila, Rosana and Branson, Megan and Davis, Kelly and Henretty, Michael and Kohler, Michael and Meyer, Josh and Morais, Reuben and Saunders, Lindsay and Tyers, Francis M. and Weber, Gregor},
  year = {2020},
  eprint = {1912.06670},
  primaryclass = {cs},
  urldate = {2022-06-15},
  archiveprefix = {arXiv}
}

@inproceedings{schuurmanCGNAnnotatedCorpus2003,
    title = "{CGN}, an annotated corpus of spoken {D}utch",
    author = "Schuurman, Ineke  and
      Schouppe, Machteld  and
      Hoekstra, Heleen  and
      van der Wouden, Ton",
    booktitle = "Proceedings of 4th International Workshop on Linguistically Interpreted Corpora ({LINC}-03) at {EACL} 2003",
    year = "2003",
    url = "https://aclanthology.org/W03-2414/"
}

@inproceedings{ottFairseqFastExtensible2019,
  title = {Fairseq: {{A Fast}}, {{Extensible Toolkit}} for {{Sequence Modeling}}},
  shorttitle = {Fairseq},
  author = {Ott, Myle and Edunov, Sergey and Baevski, Alexei and Fan, Angela and Gross, Sam and Ng, Nathan and Grangier, David and Auli, Michael},
  year = {2019},
  pages = {48--53},
  publisher = {ACL},
  address = {Minneapolis, Minnesota},
  doi = {10.18653/v1/N19-4009},
  urldate = {2022-06-15}
}

@inproceedings{pratapMLSLargeScaleMultilingual2020,
  title = {{{MLS}}: {{A Large-Scale Multilingual Dataset}} for {{Speech Research}}},
  shorttitle = {{{MLS}}},
  booktitle = {Proc. Interspeech},
  author = {Pratap, Vineel and Xu, Qiantong and Sriram, Anuroop and Synnaeve, Gabriel and Collobert, Ronan},
  eprint = {2012.03411},
  pages = {2757--2761},
  year = {2020},
  doi = {10.21437/Interspeech.2020-2826}
}

@inproceedings{baevskiWav2vecFrameworkSelfSupervised2020,
	title = {wav2vec 2.0: {A} {Framework} for {Self}-{Supervised} {Learning} of {Speech} {Representations}},
	volume = {33},
	shorttitle = {wav2vec 2.0},
	url = {https://proceedings.neurips.cc/paper/2020/hash/92d1e1eb1cd6f9fba3227870bb6d7f07-Abstract.html},
	urldate = {2024-03-09},
	booktitle = {Advances in {Neural} {Information} {Processing} {Systems}},
	publisher = {Curran Associates, Inc.},
	author = {Baevski, Alexei and Zhou, Yuhao and Mohamed, Abdelrahman and Auli, Michael},
	year = {2020},
	pages = {12449--12460}
}

@article{bojanowskiEnrichingWordVectors2017a,
	title = {Enriching {Word} {Vectors} with {Subword} {Information}},
	volume = {5},
	issn = {2307-387X},
	url = {https://doi.org/10.1162/tacl_a_00051},
	doi = {10.1162/tacl_a_00051},
	urldate = {2025-03-13},
	journal = {Transactions of the Association for Computational Linguistics},
	author = {Bojanowski, Piotr and Grave, Edouard and Joulin, Armand and Mikolov, Tomas},
	month = jun,
	year = {2017},
	pages = {135--146}
}

@article{Honnibal_spaCy_Industrial-strength_Natural_2020,
author = {Honnibal, Matthew and Montani, Ines and Van Landeghem, Sofie and Boyd, Adriane},
doi = {10.5281/zenodo.1212303},
title = {{spaCy: Industrial-strength Natural Language Processing in Python}},
year = {2020}
}

@inproceedings{shen-etal-2024-encoding,
  title = {Encoding of Lexical Tone in Self-Supervised Models of Spoken Language},
  booktitle = {Proceedings of the 2024 Conference of the North American Chapter of the Association for Computational Linguistics: {{Human}} Language Technologies (Volume 1: {{Long}} Papers)},
  author = {Shen, Gaofei and Watkins, Michaela and Alishahi, Afra and Bisazza, Arianna and Chrupa{\l}a, Grzegorz},
  editor = {Duh, Kevin and Gomez, Helena and Bethard, Steven},
  year = {2024},
  month = jun,
  pages = {4250--4261},
  publisher = {Association for Computational Linguistics},
  address = {Mexico City, Mexico},
  doi = {10.18653/v1/2024.naacl-long.239}
}

@article{brysbaertNormsAgeAcquisition2014,
	title = {Norms of age of acquisition and concreteness for 30,000 {Dutch} words},
	volume = {150},
	issn = {0001-6918},
	url = {https://www.sciencedirect.com/science/article/pii/S0001691814000985},
	doi = {10.1016/j.actpsy.2014.04.010},
	urldate = {2025-03-13},
	journal = {Acta Psychologica},
	author = {Brysbaert, Marc and Stevens, Michaël and De Deyne, Simon and Voorspoels, Wouter and Storms, Gert},
	month = jul,
	year = {2014},
	pages = {80--84}
}

@inproceedings{dugonjicWhatHasLeBenchmark2024,
    title = "What Has {L}e{B}enchmark Learnt about {F}rench Syntax?",
    author = "Dugonji{\'c}, Zdravko  and
      Pupier, Adrien  and
      Lecouteux, Benjamin  and
      Coavoux, Maximin",
    editor = "Calzolari, Nicoletta  and
      Kan, Min-Yen  and
      Hoste, Veronique  and
      Lenci, Alessandro  and
      Sakti, Sakriani  and
      Xue, Nianwen",
    booktitle = "Proceedings of the 2024 Joint International Conference on Computational Linguistics, Language Resources and Evaluation (LREC-COLING 2024)",
    month = may,
    year = "2024",
    address = "Torino, Italia",
    publisher = "ELRA and ICCL",
    url = "https://aclanthology.org/2024.lrec-main.1521/",
    pages = "17493--17499"
}

@article{baayenCELEXLexicalDatabase1996,
	title = {The {CELEX} {Lexical} {Database}},
	url = {https://pure.mpg.de/pubman/faces/ViewItemOverviewPage.jsp?itemId=item_2339741},
	language = {eng},
	urldate = {2025-03-13},
	author = {Baayen, R. H. and Piepenbrock, R. and Gulikers, L.},
	year = {1996},
	note = {Publisher: University of Pennsylvania},
}

@phdthesis{schatzABXDiscriminabilityMeasuresApplications2016,
	type = {Ph{D} dissertation},
	title = {{ABX}-{Discriminability} {Measures} and {Applications}},
	url = {https://hal.science/tel-01407461},
	language = {en},
	urldate = {2025-03-12},
	school = {Université Paris 6 (UPMC)},
	author = {Schatz, Thomas},
	month = sep,
	year = {2016}
}

@article{kriegeskorteRepresentationalSimilarityAnalysis2008a,
	title = {Representational similarity analysis - connecting the branches of systems neuroscience},
	volume = {2},
	issn = {1662-5137},
	url = {https://www.frontiersin.org/articles/10.3389/neuro.06.004.2008/full},
	doi = {10.3389/neuro.06.004.2008},
	language = {English},
	number = {4},
	urldate = {2020-01-23},
	journal = {Frontiers in Systems Neuroscience},
	author = {Kriegeskorte, Nikolaus and Mur, Marieke and Bandettini, Peter A.},
	year = {2008},
	pages = {1--28}
}

@inproceedings{hewittStructuralProbeFinding2019,
	address = {Minneapolis, Minnesota},
	title = {A {Structural} {Probe} for {Finding} {Syntax} in {Word} {Representations}},
	url = {https://aclanthology.org/N19-1419},
	doi = {10.18653/v1/N19-1419},
	booktitle = {Proceedings of the 2019 {Conference} of the {North} {American} {Chapter} of the {Association} for {Computational} {Linguistics}: {Human} {Language} {Technologies}, {Volume} 1 ({Long} and {Short} {Papers})},
	publisher = {Association for Computational Linguistics},
	author = {Hewitt, John and Manning, Christopher D.},
	month = jun,
	year = {2019},
	pages = {4129--4138}
}

@article{orhanDetectionAlgebraicAuditory2025,
	title = {The detection of algebraic auditory structures emerges with self-supervised learning},
	volume = {21},
	issn = {1553-7358},
	url = {https://journals.plos.org/ploscompbiol/article?id=10.1371/journal.pcbi.1013271},
	doi = {10.1371/journal.pcbi.1013271},
	language = {en},
	number = {9},
	urldate = {2026-03-30},
	journal = {PLOS Computational Biology},
	publisher = {Public Library of Science},
	author = {Orhan, Pierre and Boubenec, Yves and King, Jean-Rémi},
	month = sep,
	year = {2025},
	pages = {e1013271}
}

@article{changWordAcquisitionNeural2022,
	title = {Word {Acquisition} in {Neural} {Language} {Models}},
	volume = {10},
	issn = {2307-387X},
	url = {https://doi.org/10.1162/tacl_a_00444},
	doi = {10.1162/tacl_a_00444},
	journal = {Transactions of the Association for Computational Linguistics},
	author = {Chang, Tyler A. and Bergen, Benjamin K.},
	month = jan,
	year = {2022},
	pages = {1--16}
}

@article{khorramiComputationalInsightsAcquisition2023,
	title = {Computational {Insights} to {Acquisition} of {Phonemes}, {Words}, and {Word} {Meanings} in {Early} {Language}: {Sequential} or {Parallel} {Acquisition}?},
	volume = {45},
	shorttitle = {Computational {Insights} to {Acquisition} of {Phonemes}, {Words}, and {Word} {Meanings} in {Early} {Language}},
	url = {https://escholarship.org/uc/item/79t028n8},
	language = {en},
	number = {45},
	journal = {Proceedings of the Annual Meeting of the Cognitive Science Society},
	author = {Khorrami, Khazar and Cruz Blandón, María Andrea and Räsänen, Okko},
	year = {2023}
}

@inproceedings{evanson-etal-2023-language,
    title = "Language acquisition: do children and language models follow similar learning stages?",
    author = "Evanson, Linnea  and
      Lakretz, Yair  and
      King, Jean R{\'e}mi",
    editor = "Rogers, Anna  and
      Boyd-Graber, Jordan  and
      Okazaki, Naoaki",
    booktitle = "Findings of the Association for Computational Linguistics: ACL 2023",
    month = jul,
    year = "2023",
    address = "Toronto, Canada",
    publisher = "Association for Computational Linguistics",
    url = "https://aclanthology.org/2023.findings-acl.773/",
    doi = "10.18653/v1/2023.findings-acl.773",
    pages = "12205--12218"
}

@incollection{rumelhart1986learning,
  author      = {Rumelhart, D. E. and McClelland, J. L.},
  title       = {On {Learning} the {Past} {Tenses} of {English} {Verbs}},
  editor      = "McClelland, J.L. and Rumelhart, D.E. and PDP Research Group",
  url = {https://doi.org/10.7551/mitpress/5237.003.0008},
  booktitle   = "Parallel Distributed Processing, Volume 2: Explorations in the Microstructure of Cognition: Psychological and Biological Models",
  publisher   = "The MIT Press",
  year        = 1986,
  chapter     = 18,
}

@inproceedings{milletSelfsupervisedSpeechModels2022,
	address = {Dublin, Ireland},
	title = {Do self-supervised speech models develop human-like perception biases?},
	url = {https://aclanthology.org/2022.acl-long.523},
	doi = {10.18653/v1/2022.acl-long.523},
	urldate = {2024-10-10},
	booktitle = {Proceedings of the 60th {Annual} {Meeting} of the {Association} for {Computational} {Linguistics} ({Volume} 1: {Long} {Papers})},
	publisher = {Association for Computational Linguistics},
	author = {Millet, Juliette and Dunbar, Ewan},
	editor = {Muresan, Smaranda and Nakov, Preslav and Villavicencio, Aline},
	month = may,
	year = {2022},
	pages = {7591--7605}
}

@inproceedings{chrupala-etal-2020-analyzing,
    title = "Analyzing analytical methods: The case of phonology in neural models of spoken language",
    author = "Chrupa{\l}a, Grzegorz  and
      Higy, Bertrand  and
      Alishahi, Afra",
    editor = "Jurafsky, Dan  and
      Chai, Joyce  and
      Schluter, Natalie  and
      Tetreault, Joel",
    booktitle = "Proceedings of the 58th Annual Meeting of the Association for Computational Linguistics",
    month = jul,
    year = "2020",
    address = "Online",
    publisher = "Association for Computational Linguistics",
    url = "https://aclanthology.org/2020.acl-main.381",
    doi = "10.18653/v1/2020.acl-main.381",
    pages = "4146--4156",
}

@inproceedings{shenWaveSyntaxProbing2023a,
	title = {Wave to {Syntax}: {Probing} spoken language models for syntax},
	shorttitle = {Wave to {Syntax}},
	url = {https://www.isca-archive.org/interspeech_2023/shen23_interspeech.html},
        booktitle = {Proc. Interspeech},
	doi = {10.21437/Interspeech.2023-679},
	urldate = {2024-10-10},
	author = {Shen, Gaofei and Alishahi, Afra and Bisazza, Arianna and Chrupała, Grzegorz},
	year = {2023},
	pages = {1259--1263}
}

@article{magnuson2018interaction,
  title={Interaction in spoken word recognition models: Feedback helps},
  author={Magnuson, James S and Mirman, Daniel and Luthra, Sahil and Strauss, Ted and Harris, Harlan D},
  journal={Frontiers in psychology},
  volume={9},
  pages={369},
  year={2018},
  url={https://doi.org/10.3389/fpsyg.2018.00369},
  publisher={Frontiers Media SA}
}

@article{norris2016prediction,
  title={Prediction, Bayesian inference and feedback in speech recognition},
  author={Norris, Dennis and McQueen, James M and Cutler, Anne},
  journal={Language, cognition and neuroscience},
  volume={31},
  number={1},
  pages={4--18},
  year={2016},
  url={https://doi.org/10.1080/23273798.2015.1081703},
  publisher={Taylor \& Francis}
}
\bibliographystyle{acl_natbib}

\onecolumn

\appendix

\setcounter{figure}{0}
\renewcommand{\thefigure}{S\arabic{figure}}
\renewcommand{\thetable}{S\arabic{table}}

\section{Model validation}
\label{app:model-validation}
\begin{figure*}[h]
\includegraphics[width=\textwidth]{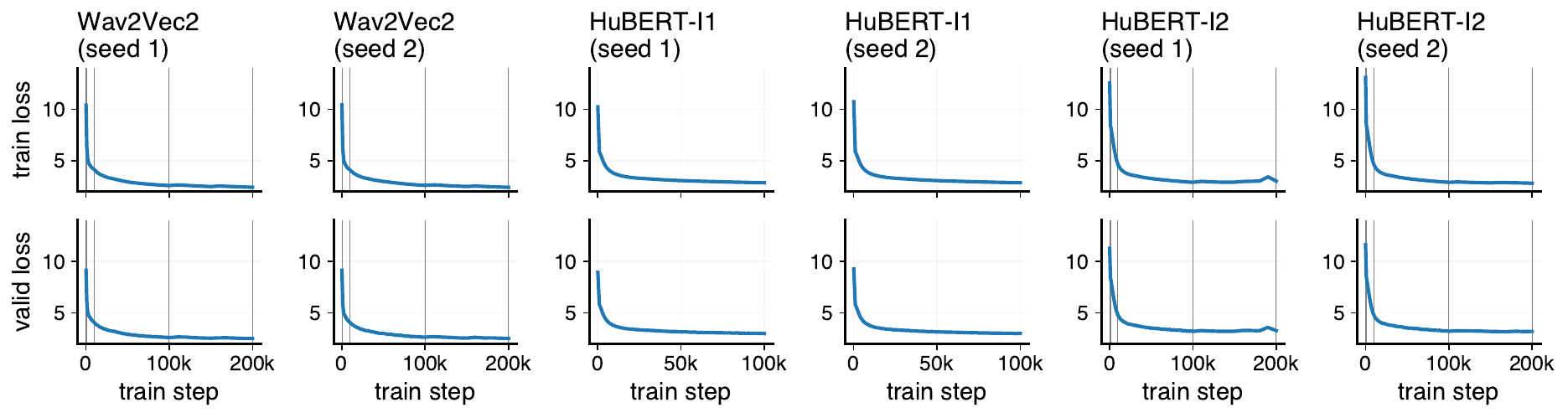}
\caption{Pre-training loss curves (train \& validation loss) for all models trained for this study. Checkpoints additionally validated for their downstream performance on automatic speech recognition (see \autoref{tab:wer-results}) are marked with grey vertical lines.}
\label{fig:loss-curves}
\end{figure*}

We validated that capable speech SSL models were trained by observing that the pre-training losses for each model showed a stable decrease over training (train and validation loss curves are visualized in \autoref{fig:loss-curves}). Additionally, we finetuned a sample of intermediate checkpoints of the Wav2Vec2 and HuBERT-I2 models on Dutch automatic speech recognition (\autoref{tab:wer-results}), confirming that word-error rates are consistently lower for checkpoints later in training, and validating that the trained models also achieve reasonable downstream task performance. We note that results presented in our main text are obtained on checkpoints pre-trained only using SSL objectives, and do not include models fine-tuned for ASR. In further training up to 200K checkpoints, we noted instabilities and degraded downstream performance for HuBERT-I2 seed 1; we therefore exclude this model from our analyses on the 200K checkpoints (\autoref{fig:200k-comp}).

\renewcommand{\cellalign}{tl}
\begin{table*}[h]
\centering
\begin{tabular}{p{3cm}|c|c|c|c|c}
\hline
Test set & Step & HuBERT-I2 (1) & HuBERT-I2 (2) & Wav2Vec2 (1) & Wav2Vec2 (2) \\
\hline
IFADV   & 100    & 0.97 & 0.98 & 0.97 & 0.97 \\
(dialogues)       & 1000   & 0.98 & 0.97 & 0.98 & 0.98 \\
       & 10000  & 0.86 & 0.86 & 0.91 & 0.91 \\
       & 100000 & \textbf{0.67} & \textbf{0.65} & 0.71 & 0.70 \\
       & 200000 & 0.68 & 0.66 & \textbf{0.69} & \textbf{0.69} \\
\hline\hline
NBest   & 100    & 0.83 & 0.85 & 0.83 & 0.82 \\
(benchmark)       & 1000   & 0.83 & 0.83 & 0.84 & 0.85 \\
       & 10000  & 0.62 & 0.60 & 0.68 & 0.69 \\
       & 100000 & \textbf{0.26} & \textbf{0.25} & 0.32 & 0.32 \\
       & 200000 & 0.27 & 0.26 & \textbf{0.30} & \textbf{0.29} \\
\hline\hline
CV    & 100    & 0.85 & 0.86 & 0.86 & 0.85 \\
(read aloud         & 1000   & 0.85 & 0.85 & 0.89 & 0.89 \\
sentences)       & 10000  & 0.65 & 0.62 & 0.72 & 0.72 \\
       & 100000 & \textbf{0.22} & 0.22 & 0.30 & 0.30 \\
       & 200000 & 0.23 & \textbf{0.21} & \textbf{0.28} & \textbf{0.27} \\
\hline\hline
MLS    & 100    & 0.67 & 0.68 & 0.67 & 0.67 \\
(audiobooks)       & 1000   & 0.68 & 0.68 & 0.70 & 0.72 \\
       & 10000  & 0.40 & 0.39 & 0.48 & 0.48 \\
       & 100000 & \textbf{0.15} & 0.15 & 0.19 & 0.20 \\
       & 200000 & \textbf{0.15} & \textbf{0.14} & \textbf{0.18} & \textbf{0.18} \\       
\hline\hline
CGN-O      & 100    & 0.52 & 0.53 & 0.51 & 0.51 \\
(audiobooks)       & 1000   & 0.52 & 0.52 & 0.54 & 0.58 \\
       & 10000  & 0.26 & 0.25 & 0.32 & 0.33 \\
       & 100000 & \textbf{0.08} & 0.08 & 0.11 & 0.11 \\
       & 200000 & 0.09 & \textbf{0.07} & \textbf{0.10} & \textbf{0.10} \\
\hline
\end{tabular}
\caption{Word-error rate (WER) results for the HuBERT-I2 (1 \& 2) and Wav2Vec2-NL (1 \& 2) models pre-trained for this study, after finetuning intermediate checkpoints for Dutch automatic speech recognition (using the CGN-O component from the Spoken Dutch corpus, \citealp{schuurmanCGNAnnotatedCorpus2003}); evaluated across several test sets: IFADV \citep{vansonIFADVCorpusFree2008}; the NBest benchmark \citep{vaessenSelfsupervisedLearningSpeech2025}; Dutch CommonVoice \citep{ardilaCommonVoiceMassivelyMultilingual2020}; Multilingual LibriSpeech \citep{pratapMLSLargeScaleMultilingual2020}; and a held-out part of CGN-O. Step refers to the pre-training step. Bold values mark the lowest WER per model and test set.}
\label{tab:wer-results}
\end{table*}

\clearpage

\section{Representational probing details}
\label{app:probe-details}

\subsection{Supplementary results}
\label{app:supp-probe-results}

\begin{figure*}[h]
    \includegraphics[width=\textwidth]{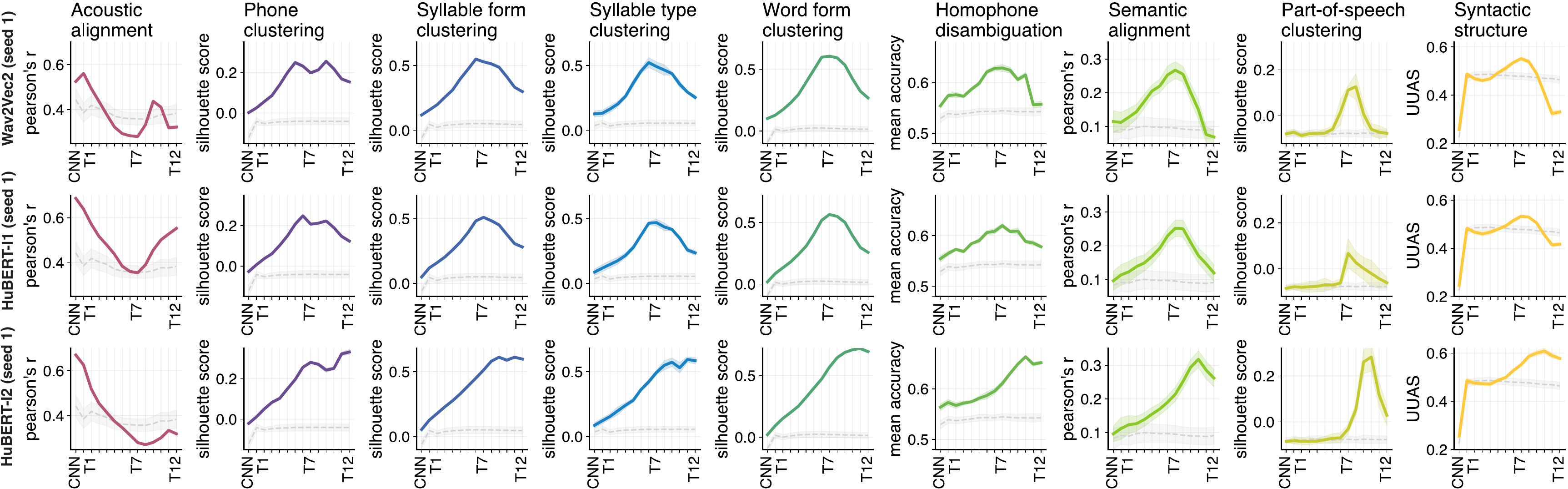}
    \caption{Layerwise scores for all representational probes (columns) and three speech SSL models (rows). Grey dashed lines indicate baseline scores extracted from a Wav2Vec2 model trained on non-speech acoustic scenes. Shading indicates the std. dev. over 5 folds. Note that scores are on different scales across probes and displayed together only to demonstrate the variety in relative layerwise peak locations, not absolute performance.}
    \label{fig:layerwise-seed1}
\end{figure*}

\begin{figure*}[h]
    \includegraphics[width=\textwidth]{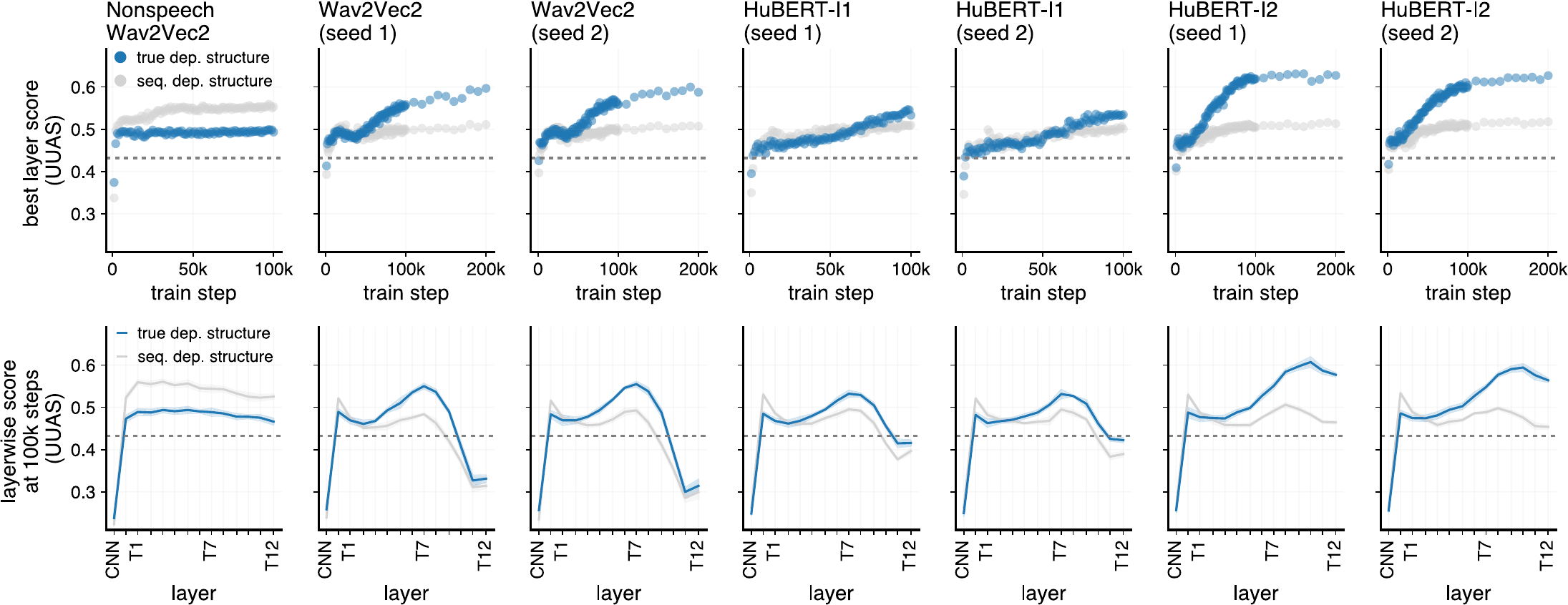}
    \caption{Control analyses for the syntactic structure probe: grey markers show alignment between model-reconstructed dependency parses and linear-sequential dependency structures (where each word is only connected to its neighbors); blue markers show alignment to the sentences' true dependency structures. In all speech-trained models, alignment to the true dependency parse is (eventually) higher than alignment to the sequential parse. This pattern is reversed for the model trained on non-speech acoustic scenes. Top row: best-layer scores across training checkpoints. Bottom row: layerwise scores for the checkpoint at 100k training steps. The dashed line indicates the mean UUAS score quantifying similarity between the true and sequential dependency structures themselves.}
    \label{fig:syntactic-control-results}
\end{figure*}

\clearpage

\newcolumntype{L}{>{\raggedright\arraybackslash}m{3cm}} 
\newcolumntype{T}{>{\raggedright\arraybackslash}m{1.35cm}} 
\newcolumntype{S}{>{\raggedright\arraybackslash}m{3.15cm}} 

\subsection{Probe dataset design}
\label{app:probe-design}

We here include details on the data used for all our probing analyses, including the number of samples (occurrences of each item sampled from MLS data) and other relevant design choices.

\begin{table}[H]
\centering
\begin{tabular}{@{}lLLTS@{}}
\toprule
\textbf{Analysis}         & \textbf{Number of categories}        & \textbf{Data samples per category}                             & \textbf{Total samples} & \textbf{Data split constraints}                                                        \\ \midrule
Phone clustering & 37 phones \hspace{3em} (see \autoref{tab:phone-set})                   & 50                                                    & 1850          & 80/20\% train-test splits                                                       \\ \midrule
Syllable form clustering    & 100 syllable forms \hspace{3em} (see \autoref{tab:syllable-set})          & 20                                                    & 2000          & 80/20\% train-test splits                                                       \\ \midrule
Syllable type clustering    & 20 types \hspace{3em} (see \autoref{tab:syllable-set})                    & 100 (5 syllable forms per type, with 20 samples each)          & 2000          & 80/20\% train-test splits, syllable forms do not overlap between train \& test \\ \midrule
Word form clustering        & 250 word forms              & 10                                                    & 2500          & 80/20\% train-test splits                                                       \\ \midrule
Part-of-speech clustering   & 4 part-of-speech categories & 150 (50 word forms per category, with 3 samples each) & 600           & 80/20\% train-test splits, word forms do not overlap between train \& test     \\ \bottomrule
\end{tabular}
\caption{Dataset details for all clustering probes. For each analysis with number of categories $N$, we train LDA projections to an $N-1$ dimensional subspace (following default configurations in the scikit-learn \citep{scikit-learn} library); i.e. phone LDA projections consist of 36 discriminative directions, etc.}
\label{tab:clustering-details}
\end{table}

\newcolumntype{X}{>{\raggedright\arraybackslash}m{8cm}} 

\begin{table}[H]
\begin{tabular}{@{}lX@{}}
\toprule
\textbf{Summary statistics: Word form analysis data} &                                                          \\ \midrule
Total word forms                     & 250                                                                                   \\ \midrule
Number of samples per word form      & 10                                                                                    \\ \midrule
Total samples                        & 2500                                                                                  \\ \midrule
Numer of word forms per POS category & 109 nouns, 44 adverbs, 25 adjectives, 21 verbs, 17 adpositions, 15 pronouns, 19 other \\ \midrule
Duration (ms)                        & Mean: 538, Std.dev: 220                                                               \\ \midrule
Length (number of phones)            & Mean: 5.94, Std.dev: 2.22                                                             \\ \midrule
Zipf frequency \citep{keuleersSUBTLEXNLNewMeasure2010}   & Mean: 3.95, Std.dev: 1.51                                                             \\ \midrule
Age of acquisition rating \citep{brysbaertNormsAgeAcquisition2014}    & Mean: 9.93, Std.dev: 2.79                                                             \\ \bottomrule
\end{tabular}
\caption{Summary statistics on words used for word form clustering analyses. For our semantic alignment analyses, we randomly sampled 60 nouns, 20 adverbs, 20 adjectives, and 20 verbs out of the available word forms in this dataset, including all samples for each form (1000 samples in total).}
\label{tab:word-data-stats}
\end{table}

\newcolumntype{X}{>{\raggedright\arraybackslash}m{7.9cm}} 

\begin{table}[]
\centering
\begin{tabular}{@{}lX@{}}
\toprule
\textbf{Summary statistics: Homophone analysis data} & \textbf{}                 \\ \midrule
Total number of triplets (A, B, X)                   & 2326                      \\
Total number of unique words                         & 4104                      \\
Word duration (ms)                                   & Mean: 591, Std.dev.: 210  \\
Word length (number of phones)                       & Mean: 6.51, Std.dev: 2.02 \\ \bottomrule
\end{tabular}
\caption{Summary statistics on words used for the homophone disambiguation analyses.}
\label{tab:homophone-data-stats}
\end{table}

\newcolumntype{X}{>{\raggedright\arraybackslash}m{7.5cm}} 

\begin{table}[]
\centering
\begin{tabular}{@{}lX@{}}
\toprule
\textbf{Summary statistics: Part-of-speech analysis data} &                                               \\ \midrule
Number of word forms per POS category                     & 50 nouns, 50 adverbs, 50 adjectives, 50 verbs \\
Number of samples per word form                           & 3                                             \\
Total samples                                             & 600                                           \\
Duration (ms)                                             & Mean: 434, Std.dev: 165                       \\
Length (number of phones)                                 & Mean: 4.85, Std.dev: 1.79                     \\
Zipf frequency \citep{keuleersSUBTLEXNLNewMeasure2010}    & Mean: 5.01, Std.dev: 0.91                     \\ \bottomrule
\end{tabular}
\caption{Summary statistics on words used for the part-of-speech clustering analyses. The words used for part-of-speech clustering are sampled from the sentence dataset used for the syntactic structural probe (\autoref{tab:sentence-data-stats})}
\label{tab:pos-data-stats}
\end{table}

\newcolumntype{X}{>{\raggedright\arraybackslash}m{6.8cm}} 
\begin{table}[]
\centering
\begin{tabular}{@{}lX@{}}
\toprule
\textbf{Summary statistics: Sentence analysis data}   &                           \\ \midrule
Total number of sentences                             & 2406                      \\
Number of audiobooks that sentences were sourced from & 10                        \\
Sentence duration (s)                                 & Mean: 4.15, Std.dev: 2.45 \\
Sentence length (number of words)                     & Mean: 10.6, Std.dev: 5.13 \\
Word duration (ms)                                    & Mean: 325, Std.dev: 188   \\
Word dependency depth                                 & Mean: 1.73, Std.dev: 1.07 \\ \bottomrule
\end{tabular}
\caption{Summary statistics on sentences used for the syntactic structural probe analyses.}
\label{tab:sentence-data-stats}
\end{table}

\newcolumntype{F}{>{\raggedright\arraybackslash}m{3.5cm}} 
\newcolumntype{X}{>{\raggedright\arraybackslash}m{8.6cm}} 

\begin{table}[]
\begin{tabular}{@{}lFX@{}}
\toprule
\textbf{Analysis}  & \textbf{Feature space}                      & \textbf{Feature details}                                                                                                   \\ \midrule
Acoustic alignment & Mel-frequency cepstral coefficients (MFCCs) & 20-dimensional, extracted using the librosa library \citep{mcfeeLibrosaAudioMusic2015}, mean-pooled over time                                                 \\ \midrule
Semantic alignment & Fasttext embeddings                         & 300-dimensional, Fasttext model \citep{bojanowskiEnrichingWordVectors2017a} for Dutch released by \citet{mikolov2018advances}, accessed from the HuggingFace Hub through the transformers library \citep{wolfHuggingFacesTransformersStateoftheart2020} \\ \bottomrule
\end{tabular}
\caption{Details on the feature spaces used for the acoustic and semantic alignment analyses. Acoustic alignment was computed on the same set of 1850 samples used for phone clustering (see \autoref{tab:clustering-details}).}
\label{tab:rsa-features}
\end{table}

\begin{table}[h]
\resizebox{0.5\columnwidth}{!}{
\begin{tabular}{@{}lll@{}}
\toprule
\textbf{\begin{tabular}[c]{@{}l@{}}Broad phon.\\ category\end{tabular}} &
  \textbf{\begin{tabular}[c]{@{}l@{}}Fine phon.\\ category\end{tabular}} &
  \textbf{\begin{tabular}[c]{@{}l@{}}Included \\ phones\end{tabular}} \\ \midrule
consonant & plosive     & \textipa{p, b, t, d, k, g}            \\
          & fricative   & \textipa{f, v, s, z, S, x, G, h}      \\
          & nasal       & \textipa{m, n, N}                     \\
          & approximant & \textipa{V, l, r, j}                  \\
vowel     & long        & \textipa{i:, y:, e:, \o:, a:, o:, u:} \\
          & short       & \textipa{I, E, A, O, 0}               \\
          & diphthong   & \textipa{Ei, \oe y, Au}               \\
          & reduced     & \textipa{@}                           \\ \bottomrule
\end{tabular}}
\caption{Dutch phones included in our phone-level \\\hspace{\textwidth} analyses, along with their broad and fine category.}
\label{tab:phone-set}
\end{table}

\begin{table}[h]
\resizebox{0.5\textwidth}{!}{%
\begin{tabular}{@{}ll@{}}
\toprule
\textbf{Syllable type} & \textbf{Syllable forms}                          \\ \midrule
CV                     & \textipa{d@, t@, x@, b@, l@}                     \\
CVC                    & \textipa{v@r, d@r, l@k, k@r, t@x}                \\
CVV                    & \textipa{na:, tu:, ma:, re:, le:}                \\
CVVC                   & \textipa{hEit, vo:r, Va:r, Ve:r, h\oe ys}        \\
VC                     & \textipa{Op, Af, Om, In, Ax}                     \\
CVCC                   & \textipa{k@nt, d@rs, r@nt, V@rk, G@nt}           \\
VVC                    & \textipa{\oe yt, a:n, e:n, o:r, e:r}             \\
CCVV                   & \textipa{tsi:, ste:, dra:, sxa:, pro:}           \\
CVVCC                  & \textipa{vo:rt, te:rt, ho:ft, ma:kt, be:lt}      \\
CCVVC                  & \textipa{sta:t, sxo:l, sla:n, kle:t, pla:t}      \\
VV                     & \textipa{o:, a:, e:, i:, y:}                     \\
CCVC                   & \textipa{tj@s, st@r, sxAp, slAx, stOr}           \\
CCV                    & \textipa{st@, tj@, pj@, pl@, br@}                \\
VCC                    & \textipa{Ont, Axt, Ant, Arm, Eks}                \\
CCVCC                  & \textipa{stElt, st@rs, stAnt, brAxt, stOrm}      \\
CCVVCC                 & \textipa{pla:ts, krEixs, sta:ts, bre:kt, sxe:ps} \\
CCCVV                  & \textipa{sxrei, spre:, stra:, sxre:, strEi}      \\
CCCVC                  & \textipa{strAf, sprON, sxrIf, sxrOm, strEk}      \\
CVVCCC                 & \textipa{di:nst, Va:rts, ka:tst, la:tst, vEinst} \\
CCCVCC                 & \textipa{sprINt, sxrIkt, strEkt, strAft, strAnt} \\ \bottomrule
\end{tabular}%
}
\caption{Included syllable forms for each syllable type, \\\hspace{\textwidth}ordered by frequency of syllable type as estimated from \\\hspace{\textwidth}the Spoken Dutch Corpus \citep{zuidemaSyllableFrequencyList}. We sample \\\hspace{\textwidth}all syllable forms from multi-syllabic words to avoid \\\hspace{\textwidth}confounding effects of word form clustering on the \\\hspace{\textwidth}syllable-level analyses.}
\label{tab:syllable-set}
\end{table}

\end{document}